\documentclass[10pt,twocolumn,letterpaper]{article}

\usepackage[pagenumbers]{cvpr} 

\newcommand{\JA}[1]{{\color{blue}\textbf{JA:#1}}}

\newcommand{\TD}[1]{{\color{red}\textbf{TD:#1}}}

\usepackage{graphicx}
\usepackage{float}
\usepackage{algorithm}
\usepackage{algpseudocode}
\usepackage{booktabs}
\usepackage{multirow}

\newcommand{\set}[1]{\mathcal{#1}}
\newcommand{\dataset}[1]{\mathcal{D}}
\newcommand{\trainset}{\mathcal{D}_{train}}
\newcommand{\valset}{\mathcal{D}_{val}}

\newcommand{\poolset}{\mathcal{D}_{pool}}
\newcommand{\importance}{\mathcal{I}}
\newcommand{\navtrain}{\texttt{Navtrain}}
\newcommand{\openscene}{\texttt{Openscene}}

\newcommand{\tss}[1]{\textsuperscript{#1}}

\newcommand{\method}{\textit{MOSAIC}}
\newcommand{\methodlong}{Mixture Optimization via Scaling-Aware Iterative Collection}

\definecolor{cvprblue}{rgb}{0.21,0.49,0.74}
\usepackage[pagebackref,breaklinks,colorlinks,allcolors=cvprblue]{hyperref}

\usepackage{amsmath,amssymb}   
\usepackage{algorithm}    
\usepackage{algpseudocode}       

\def\paperID{REMOVED} 
\def\confName{CVPR}
\def\confYear{2026}

\title{ Scaling-Aware Data Selection for End-to-End Autonomous Driving Systems }

\author{Tolga Dimlioglu\tss{1}\footnotemark[2] , Nadine Chang\tss{2}, Maying Shen\tss{2}, Rafid Mahmood\tss{2,3}, Jose M. Alvarez\tss{2} \\
\tss{1}New York University, \tss{2}NVIDIA, \tss{3}University of Ottawa\\
{\tt\small td2249@nyu.edu, \{nadinec, mshen, rmahmood, josea\}@nvidia.com}
}

\begin{document}
\maketitle
\begin{abstract}
Large-scale deep learning models for physical AI applications depend on diverse training data collection efforts. These models and correspondingly, the training data, must address different evaluation criteria necessary for the models to be deployable in real-world environments. 
Data selection policies can guide the development of the training set, but current frameworks do not account for the ambiguity in how data points affect different metrics. 
In this work, we propose Mixture Optimization via Scaling-Aware Iterative Collection (MOSAIC), a general data selection framework that operates by: (i) partitioning the dataset into domains; (ii) fitting neural scaling laws from each data domain to the evaluation metrics; and (iii) optimizing a data mixture by iteratively adding data from domains that maximize the change in metrics. 
We apply MOSAIC to autonomous driving (AD), where an End-to-End (E2E) planner model is evaluated on the Extended Predictive Driver Model Score (EPDMS), an aggregate of driving rule compliance metrics. 
Here, MOSAIC outperforms a diverse set of baselines on EPDMS with up to 80\% less data. 
%
%
%
\end{abstract}    
\footnotetext[2]{work done during internship at NVIDIA.}

\vspace{-1mm}
\section{Introduction}
\label{sec:intro}


Large-scale deep learning models are fueled by diverse data collection efforts \citep{pattnayak2024survey, liu2024survey}. This practice is particularly prominent in physical artificial intelligence (AI) applications such as autonomous driving (AD), where video clips are collected over different locations, weather, and traffic conditions \citep{Grzywaczewski2017NvidiaScale, Coren2025TeslaData}. 
It is computationally inefficient to train models on all collected data, which in physical AI can scale to hundreds of millions of hours of clips. 
This necessitates data mixture selection policies to construct and grow training sets of diverse and influential samples that maximize desired performance metrics.

\begin{figure}
    \centering
    \includegraphics[width=1.0\linewidth]{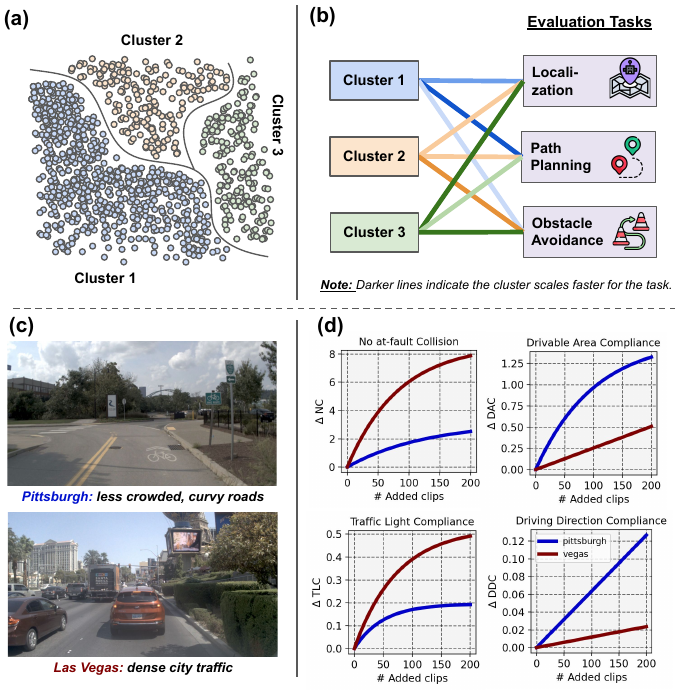}
    \caption{(a-b) The data pool is partitioned into a set of discrete domains which may each contribute to performance improvement of different evaluation tasks at varying rates. (c-d) Example application in autonomous driving: two clusters representing different driving contexts—Pittsburgh (curvy suburban roads) and Las Vegas (dense urban traffic). Data from separate contexts influence different rule-compliance metrics at distinct rates.}
    \label{fig:teaser}
    \vspace{-9mm}
\end{figure}

Dataset selection and optimization has been broadly studied from various perspectives. 
For instance, influence (or duplicate) estimation techniques use feature information to select useful data samples \cite{abbas2023semdedup, broder1997resemblance, sorscher2022beyond}, while active learning strategies optimize over this feature space \citep{sener2017active, mahmoodlow}. 
Large language models (LLMs) and their multi-modal extensions have successfully leveraged scaling laws to forecast how model performance improves with dataset size \citep{kaplan2020scaling, henighan2020scaling, zhai2022scaling}; this premise has expanded to other applications including AD \citep{baniodeh2025scaling}.
Further, as data collection becomes increasingly complex, scaling laws are used to determine optimal mixtures of data from explicit domains (e.g., different languages, math, coding) \citep{jiang2025adaptive, ye2025data, liu2025regmix}. 
Although these methods present a general opportunity for physical AI systems, they are not immediately usable for applications that require both understanding and interacting with diverse real-world scenarios for three main reasons. 
First, physical AI systems are evaluated over a set of potentially competing metrics \cite{dauner2024navsim, yu2020meta}. Second, different data samples can influence different combinations of these metrics at various rates. Finally, the data pool is not necessarily immediately separable into subsets that have consistent, predictable influence on the metrics. Existing data mixture methods assume well-defined and homogeneous domains. However, they overlook the heterogeneous and metric-dependent improvement rates that arise when data sources influence different aspects of performance at varying rates~\cite{hagele2024scaling, xie2025chameleon}. For example, a physical AI system such as an autonomous vehicle must progress along a route, follow driving rules, and avoid collisions \citep{dauner2024navsim}. 
High-traffic and pedestrian-heavy driving clips, when used for training, may impact certain metrics more than others. 
Moreover, finding such a subset of potential training data that has shared effects on the metrics requires careful selection and mining.

In this work, we develop a data selection and mixture optimization policy that addresses the present physical AI challenges of multiple competing metrics and imprecise data partitions. To address the challenge of imprecise data domains, we first partition a data pool into a set of separable clusters, within which we can rank samples on their influence to the metrics. 
We estimate the impact of each cluster on each metric, and correspondingly, an overall utility function that aggregates all the metrics.
This impact is measured in terms of scaling laws that estimate the improvement to the metrics if more data from a specific cluster were used for training. 
Finally, we iteratively add new data to the training set by identifying the cluster which will maximize the expected gain to the aggregate utility with each additional data point. In this way, we optimize the mixture of data from our generated partitions. Figure~\ref{fig:teaser} summarizes the challenges.

We apply our framework, \methodlong~(\method), for End-to-End (E2E) autonomous driving, where the challenges of data heterogeneity and metric competition are particularly pronounced. The goal is to optimize the Extended Predictive Driving Model Score (EPDMS), which aggregates a diverse set of rule-compliance metrics.
\method~is more data efficient than existing methods and achieves better EPDMS performance than na\"ive baselines with up to 82\% less additional data. 
Our contributions are:
\begin{itemize}
    \item We propose \method, a generic data mixture optimization pipeline that (i) clusters and ranks data, (ii) models domain-specific data scaling, and (iii) mines samples to maximize the expected gain over aggregate metrics.

    \item We apply \method\ to End-to-End Autonomous Driving (E2E AD) on the NAVSIM and OpenScene benchmarks using the challenge winning Hydra-MDP model \cite{li2024hydra}, where it achieves substantially higher driving performance than existing data selection and mixture baselines, and improve data efficiency by up to 82\%. Moreover, \method\ achieves the full training performance while requiring 42\% less data samples.
    
    \item We empirically demonstrate the necessity and robustness of our joint clustering and scaling procedure. First, \method~outperforms baselines regardless of the clustering approach (e.g., semantic captions, geolocation). Second, embedding scaling laws on top of clustering significantly outperforms clustering-only strategies. This underscores the importance of our principled data selection strategy, which leverages the estimated improvement rates of different data clusters to maximize model performance under limited data budgets.

    
\end{itemize}

\section{Related Works}
\label{sec:related_works}



\noindent
\textbf{Data Mixtures.} Recent work has highlighted the importance of how data from different domains are combined for large-scale model training \cite{mahmood2022optimizing, xie2023doremi, fan2023doge, ye2025data, liu2025regmix, xu2025unveiling, xie2025chameleon, jiang2025adaptive}. DoReMi~\cite{xie2023doremi} employs two proxy models to estimate domain weights based on excess loss, which are later used to reweigh domains when training a larger model. DOGE~\cite{fan2023doge}, tracks domain-specific gradients while training the proxy model to better capture inter-domain dynamics. Chameleon~\cite{xie2025chameleon} instead leverages kernel similarity scores computed in the model’s latent space to assign adaptive weights to data from different sources. Another line of work treats data mixture optimization as a regression problem: many small proxy models are trained with varying mixtures, and a regressor is then fit to predict the optimal mixture at larger scale~\cite{liu2025regmix, ye2025data}. A particularly relevant approach to ours is ADO~\cite{jiang2025adaptive}, which begins with a random data mixture and fits scaling estimators on the fly during training. The gradients of these estimators are used for mixture reweighting. However, ADO does not model how performance scales with different data sources in isolation, and it requires a temporal averaging mechanism with multiple hyperparameters to maintain the precision of scaling fits. Although the aforementioned data mixture methods assign weights to samples from different domains, these weights can also be interpreted as sampling probabilities for constructing mixtures with varying domain ratios. In our experiments, we adopt Chameleon~\cite{xie2025chameleon} as a baseline, since it has been shown to outperform other mixture algorithms.

\noindent
\textbf{Data Pruning \& Selection.} Data pruning aims to identify a compact subset of training data by removing redundant samples while preserving model performance \cite{shen2025sse}. In vision tasks, \citet{abbas2023semdedup} proposed removing visually similar samples using cosine similarity in the CLIP~\cite{radford2021learning} feature space. Follow-up works extended this idea to specialized domains such as object detection~\cite{kang2025adadedup} and fairness-aware multimodal learning~\cite{slyman2024fairdedup}. It has also been shown analytically that optimal pruning strategies can improve power-law scaling behavior~\cite{sorscher2022beyond}. A closely related line of work, Active learning (AL), aims to maximize model performance improvement under a limited annotation budget \cite{ren2021survey, zhan2022comparative}. In this setting, the model has access to a large unlabeled data pool and, based on some selection signal the most informative samples are identified to be used for training. Early works focused on using the model’s prediction uncertainty, quantified through posterior probabilities~\cite{lewis1994heterogeneous}, classifier margins~\cite{roth2006margin}, or entropy~\cite{joshi2009multi}. A notable method, CoreSet~\cite{sener2017active}, seeks representation diversity by mining samples that maximize coverage in the latent space. Other data selection strategies quantify the sample importance using expensive signals such as influence on model updates~\cite{liu2021influence}, gradient-based criteria~\cite{chhabra2024data} or forgetting score \cite{toneva2018an}.



\noindent
\textbf{End-to-End Autonomous Driving.} This task aims to train planner models that map raw sensory inputs directly to control commands. Early approaches~\cite{BojarskiDelTestaEtAl2016, codevilla2018end, ALVINN} learned control actions from RGB inputs via imitation learning, while later works incorporated richer input modalities such as LiDAR and navigational commands~\cite{chitta2022transfuser, wu2022trajectory}. Recently, conventional open-loop metrics have been shown to correlate poorly with closed-loop driving quality~\cite{li2024ego, dauner2023parting}. This motivated the development of simulation benchmarks that better reflect real-world driving performance~\cite{dauner2024navsim} and, AD models to employ probabilistic and rule-compliant trajectory planners~\cite{jiang2023vad, chen2024vadv2, li2024hydra}.

\begin{figure*}[h!]
    \centering
    \includegraphics[width=1.0\linewidth]{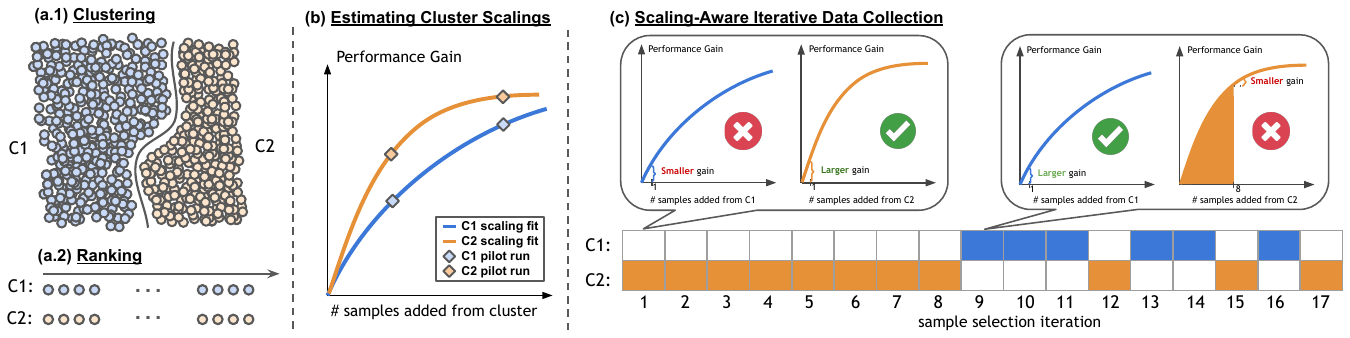}
    \vspace{-4mm}
    \caption{Overview of the proposed \textit{MOSAIC} framework. (a) The pool $\poolset$ is clustered and ranked by sample importance. (b) Cluster-wise scaling laws are fitted on pilot runs to estimate how performance scales with added data. (c) Samples are then iteratively mined from the cluster with the highest estimated marginal gain under the fitted scaling laws.}
    \label{fig:method_diagram}
    \vspace{-4mm}
\end{figure*}

\section{\method} 
\label{sec:scaling_aware_method}


\subsection{Main Problem}
\label{subsec:problem_formulation}

We want to train a Deep Neural Network (DNN) $f(\cdot; \set{D})$ on a dataset $\set{D}$ to perform a given task. 
We evaluate model performance using a set of $R$ metrics $\set{G}_r(f(\cdot; \set{D}), \valset)$ for $r \in \{1,\cdots,R\}$, where $\valset$ is a held-out validation dataset.
For brevity, we denote $\set{G}r(f(\cdot; \set{D}), \valset)$ by $\set{G}_r(\set{D})$. 
To balance the trade-offs between the metrics, we use a utility function $U(\{\set{G}_r(\set{D})\}_{r=1}^R)$ that aggregates each metric into a final score.
We do not assume about the structure of the utility function; for example, the simplest approach would be a summation $U(\cdot) = \sum_{r=1}^R \set{G}_r(\set{D})$.

We initialize with a current training dataset $\set{D}_{train}$ and a data pool $\set{D}_{pool}$. 
Given a budget $B$, our goal is to select a subset $\set{D}_{sel} \subset \set{D}_{pool}$ with $|\set{D}_{sel}| = B$ that maximizes the improvement in model performance when $f$ is retrained on the combined dataset $\set{D}_{train} \cup \set{D}_{sel}$. Formally, we write
%

\begin{equation}
\max_{\substack{\set{D}_{sel} \subset \set{D}_{pool} \\ |\set{D}_{sel}| = B}}
U \Big( \{ \set{G}_r(\set{D}_{train} \cup \set{D}_{sel}) \}_{r=1}^R \Big)
\label{eq:optim_objective}
\end{equation}

To solve problem \eqref{eq:optim_objective}, we must determine how each data sample added in $\set{D}_{sel}$ influences each of the metrics, while optimizing the trade-offs between these metrics to maximize $U(\cdot)$.
For instance, in our AD application, $f(\cdot; \set{D})$ is a planner model that maps sensory inputs to a predicted driving trajectory. 
Here, our goal is to identify driving clips that optimize the Extended Predictive Driving Model Score (EPDMS) \cite{karnchanachari2024towards, Cao2025CORL}, which is an aggregate of $R=9$ closed-loop rule compliance scores: NC, DAC, DDC, TLC, EP, TTC, LK, HC, EC. Each driving clip can showcase only certain aspects of driving rule compliance, e.g. driving on a curvy road might improve lane keeping while degrading the comfort, however our goal is to elevate the model performance across all metrics. 
Consequently, solving this problem requires disentangling the relationships between the data samples and the metrics before optimizing the trade-offs between them.


\begin{algorithm}[t]
\caption{Mixture Optimization via Scaling-Aware Iterative Selection (MOSAIC)}
\label{alg:scaling_aware}
\begin{algorithmic}[1]
\Require Pool dataset $\mathcal{D}_{pool}$, number of clusters $M$, sample selection budget $B$.
\Ensure Selected dataset $\mathcal{D}_{sel}$
\State $\{ \mathcal{D}_{pool}^{i, ranked} \}_{i=1}^M= \texttt{ClusterAndRank}(\mathcal{D}_{pool}, M)$
\State $\{\Delta \hat{U_i}(n)\}_{i=1}^M = \texttt{GetScalings}( \{ \mathcal{D}_{pool}^{i, ranked} \}_{i=1}^M )$
\State $\mathcal{D}_{sel} \gets \{\}$
\State $b_i \gets 0$ for all $i \in \{1, \dots, M\}$ 
\While{$|\mathcal{D}_{sel}| < B$}
    \For{$i = 1$ to $M$}
        \State $\delta_i(b_i) \gets\widehat{\Delta U_i}(b_i + 1) - \widehat{\Delta U_i}(b_i)$ 
    \EndFor
    \State $j \gets \arg\max_i \delta_i(b_i)$  
    \State $\text{sample} \gets \texttt{ReturnSample}(\mathcal{D}_{pool}^{j, \text{ranked}}, b_j)$ 
    \State $\mathcal{D}_{sel} \gets \mathcal{D}_{sel} \cup \{\text{sample}\}$ 
    \State $b_j \gets b_j + 1$ 
\EndWhile
\State \Return $\mathcal{D}_{sel}$
\end{algorithmic}
\end{algorithm}

\subsection{Scaling-Aware Iterative Collection} 


We propose \textit{\methodlong} (\method); a three-stage data selection framework: (i) first, cluster the pool $\set{D}_{pool}$ into partitions that capture distinct driving scene contexts and rank the data samples based on an importance score within each cluster; (ii) estimate the scaling law of adding data from each cluster with respect to the utility $U$; and (iii) iteratively mine samples from the clusters to optimize~\ref{eq:optim_objective}. Figure~\ref{fig:method_diagram} visualizes our framework. Algorithm \ref{alg:scaling_aware} summarizes the steps.

\subsubsection{Clustering \& Ranking the Data}
\label{subsubsec:clustering_n_ranking}



Before solving problem \eqref{eq:optim_objective}, we first disentangle the relationships between the metrics and data samples by clustering the data pool into a set of structured domains \cite{shen2025sse, diao2025climb}. Our goal is to find subsets of the data pool that have similar influence, i.e., samples that all influence the same set of metrics. 
Given a feature representation, we cluster the data pool into $M$ domains, i.e., $\set{D}_{pool} = \bigcup_{i=1}^M \set{D}_{pool}^{i}$. 
For example in AD, we may partition clips into clusters of highway driving, busy intersections, and calm local streets, that primarily address ego progress, collision avoidance, and traffic light compliance, respectively.



Although, clustering separates the data into domains of similar influence, each domain will include samples that have stronger influence than others \cite{katharopoulos2018not, lapedriza2013all}. When adding data, we should first exhaust the higher-influence samples \citep{sorscher2022beyond}. 
As a result, we rank the samples $x$ within each cluster via an importance score $\importance(x)$. 
In our application, we define importance by evaluating the model on that sample 
$\importance(x) := U(\{\set{G}_r(f(\cdot;\set{D}_{train}), x)\}_{r=1}^R)$. Later when adding data from each cluster, we first select samples with higher $\importance(x)$ (Line 1 of Algorithm~\ref{alg:scaling_aware}).

\subsubsection{Selecting Data by Optimizing a Mixture}
\label{subsubsec:scaling_fits}

Given a set of discrete domains, problem \eqref{eq:optim_objective} can be reformulated into a data mixture optimization problem. 
Let $\set{D}_{sel}^i \subset \set{D}_{pool}^i$ be the data added from the $i$-th domain and let $\set{D}_{sel} = \cup_{i=1}^M \set{D}_{sel}^i$. 
Furthermore, because each sample in $\set{D}_{pool}^i$ is ranked via importance scores, it remains only to determine how many samples to draw from each domain.

Mathematically, we reformulate problem \eqref{eq:optim_objective} to a proxy optimization problem below
\begin{equation}
    \max_{n_1, \cdots, n_M, \sum_{i=1}^M n_i = B} \Delta U_{mix}(n_1, \cdots, n_M) 
    \label{eq:mixture_optim}
\end{equation}
where $n_i := |\set{D}_{sel}^i|$ is the number of samples drawn from the $i$-th domain and $\Delta U_{mix}(n_1,\cdots,n_M)$ is 
\begin{equation*} 
     U \left ( \{ \set{G}_r( \trainset \cup \{ \bigcup_{i=1}^M \set{D}^{i}_{sel} \}) \}_{r=1}^R  \right ) - U( \{ \set{G}_r(\trainset)\}_{r=1}^R)
\end{equation*}
the change in utility after adding $n_i$ points from each domain. Note that the objective above is equivalent to optimizing $U ( \{ \set{G}_r( \trainset \cup \{ \cup_{i=1}^M \set{D}^{i}_{sel} \} \}_{r=1}^R ) )$, but we use the above formulation since it explicitly expresses the problem in terms of performance gains from additional data.

Solving problem \eqref{eq:mixture_optim} requires quantifying how adding data samples from each domain will improve $U(\cdot)$. We estimate this by approximating $\Delta U_{mix}$ into separate effect from each cluster and then estimating the effect of data from each domain via a scaling law. 
First, we apply the following linear separable  approximation
\vspace{-1mm}
\begin{equation}
    \Delta U_{mix}(n_1,\dots,n_M) \;\approx\; \sum_{i=1}^M \Delta U_i(n_i). 
    \label{eq:separable_approx}
\end{equation}
\vspace{-1mm}
where each $\Delta U_i(n)$ is the improvement in utility when adding only the data from the $i$-th domain
\begin{equation*} 
    U \Big ( \big \{ \set{G}_r(\set{D}_{train} \cup \set{D}_{sel}^i ) \big \}_{r=1}^R\Big ) \notag - U \Big ( \big \{ \set{G}_r(\set{D}_{train} ) \big \}_{r=1}^R\Big )
\end{equation*}

Intuitively, the approximation in \eqref{eq:separable_approx} assumes that each domain has an independent effect on the overall $\Delta U_{mix}$. 
Moreover, this assumption allows us to estimate how model performance scales if we add data from each domain independently. We use a saturating exponential scaling law
\begin{equation}
    \Delta U_i(n) \;\approx\; \widehat{\Delta U_i}(n) := a_i (1 - e^{-n / \tau_i})
    \label{eq:h_i2z}
\end{equation}
where $a_i$ and $\tau_i$ are learnable parameters of the scaling law estimated from small-scale pilot-runs (we provide details in Section~\ref{x:scaling_fits} of the Appendix.), and $n$ denotes the number of added samples \cite{xu2025unveiling}. Here, $a_{i}$ represents the asymptotic improvement on the total utility $U$ when sampling from domain $i$, while $\tau_{i}$ governs the saturation rate, i.e., how quickly the marginal benefit of adding data from the domain decreases. Obtaining the scaling esimators is symbolically captured in line 2 of Algorithm~\ref{alg:scaling_aware}. Then, substituting \eqref{eq:h_i2z} into problem \eqref{eq:mixture_optim} yields a concave maximization problem that we can compute and solve.

\subsubsection{Scaling-Aware Iterative Data Collection by First-Difference Steps}
\label{subsubsec:greedy_mining}

We propose an efficient algorithm to solve problem \eqref{eq:mixture_optim} by allocating data samples one-by-one from the domain that stands to give the highest marginal improvement to $U_{mix}$ at any given time. Intuitively, this iterative adding of data mimics a gradient-based approach of taking small steps to optimize the data mixture.

Suppose that we have so far added $b_i$ data points from each cluster in order to generate $\set{D}_{sel}$. 
Then, let
\begin{equation}
    \delta_i(b_i) \;:=\; \widehat{\Delta U_i}(b_i + 1) - \widehat{\Delta U_i}(b_i),
    \label{eq:marginal_gain}
\end{equation}
be the marginal improvement in $\widehat{\Delta U_i}(b_i)$ if we draw one additional data point from the $i$-th domain. Mathematically, $\delta_i(b_i)$ is an approximate first-difference analogue of the partial derivative of $U_{mix}$. 
Furthermore, because $\widehat{\Delta U_i}(n)$ is a concave function of $n$, this difference decreases as $b_i$ increases. This means that at a certain point, each domain yields diminishing value to the training dataset and we should draw from other domain.

In our algorithm, we iteratively add data from the domain with the highest marginal improvement. In each iteration, if we have so far drawn $b_i$ samples from the $i$-th domain, we first identify $j = \arg\max_i \delta_i(b_i)$. We then sample a data point from $\set{D}_{pool}^i$ according to the importance scores $\importance(x)$. We then update our counts $b_i$, and repeat the process until we have reached the budget (lines 5-14 in Algorithm~\ref{alg:scaling_aware}).

\section{Experiments}

We empirically evaluate \method\ on two different datasets (\openscene\ and \navtrain) using a challenge-winning model Hydra-MDP, and report consistent gains in the model performance at all budgets while being up to 80\% data efficiency than the baselines.

\subsection{Protocols}
We provide more details in Section~\ref{x:exp_protocols} of Appendix.
\vspace{-4mm}
\paragraph{Datasets.}
We use two train–pool configurations: the curated \texttt{Navtrain}~\cite{dauner2024navsim} split and the full \texttt{trainval} split of \texttt{Openscene}~\cite{contributors2023openscene}. For clarity, we refer to the latter as the \texttt{Openscene} experiment. In both settings, evaluation is conducted on the curated validation split \texttt{navtest}~\cite{dauner2024navsim}. Both datasets contain driving session clips lasting from 30 seconds to 50 minutes, which has significant temporal variation over a limited number of sessions. 
Consequently, we segment each session into fixed-length 10-second \textit{virtual clips} (20 frames at 2 Hz) and by doing so, we align our data handling practice with the industry standards~\cite{waymo_open_motion_dataset}. In the experiments, each virtual clip is treated as a single sample. 

For \texttt{Navtrain}, we use the dataset as both $\trainset$ and $\poolset$, comprising 4,601 virtual clips.
We randomly select 460 clips for $\trainset$, with the remaining 4,141 clips forming $\poolset$.
We evaluate all methods under budgets $B \in \{100, 200, 400, 800, 1600, 2400\}$. For \texttt{OpenScene}, we randomly select 1,000 clips as the $\trainset$ and reserving the remaining 31,539 as $\poolset$.
The sample selection budgets of this setting are $B \in \{250, 500, 1000, 2000, 4000, 8000\}$.

\paragraph{Model.}
We use the Hydra-MDP model~\cite{li2024hydra}, the winner of the NAVSIM Challenge in 2024 \cite{dauner2024navsim}, with a pretrained VoVNetV2-99 backbone~\cite{lee2020centermask, park2021pseudo}. The trajectory vocabulary size is set to 16,384.  For \texttt{Openscene} experiments, rule-based distillation is disabled due to the substantial pre-processing time required to compute compliance scores.


\paragraph{Baselines.}
We compare MOSAIC against several baseline data selection strategies: 
\begin{itemize}
    \item \textit{Random}: selects clips uniformly from the pool dataset under the given selection budget.
    \item \textit{Uncertainty} \cite{joshi2009multi}: measured via the entropy of the trajectory logits. Samples with higher entropy are prioritized.
    \item \textit{Coreset} \cite{sener2017active}: selects samples from the pool that maximize diversity over the feature space.
    \item \textit{Chameleon}~\cite{xie2025chameleon}: a data mixture framework that uses kernel ridge scores using domain embeddings in the model’s feature space to assign mixture weights to each domain.
\end{itemize}
Pseudo-codes are in Section~\ref{x:exp_protocols} of Appendix.
For Chameleon and \method, we cluster $\mathcal{D}_{pool}$ into domains defined by the map metadata (i.e., Boston, Pittsburgh, Singapore, Vegas). Each experiment is repeated with three random seeds. For \openscene\ experiments with more than 1,000 clips, we use two seeds to reduce computational cost. Reported results are averaged over runs, and the standard deviation is shown as a subscript in the tables.

\paragraph{Metrics.}
We evaluate models using the EPDMS, an aggregate of nine rule-compliance metrics that has been shown to correlate strongly with closed-loop driving performance~\cite{dauner2024navsim, Cao2025CORL}. Consequently, EPDMS has become the standard evaluation metric for AD planners, replacing conventional open-loop measures such as ADE and FDE. Formally, EPDMS is computed as:
\begin{align*}
    \text{EPDMS} := \prod_{m\in\mathcal{M}_{\text{pen}}}m \;.\; \frac{\sum_{m\in\mathcal{M}_{\text{avg}}}w_m m}{\sum_{m\in\mathcal{M}_{\text{avg}}}w_m} 
\end{align*}
 where $\mathcal{M}_{\text{pen}} := \{ \text{NC}, \text{DAC}, \text{DDC}, \text{TLC} \}$ denotes the set of penalty terms, and $\mathcal{M}_{\text{avg}} := \{ \text{EP}, \text{TTC}, \text{LK}, \text{HC}, \text{EC} \}$ denotes the metrics combined via a weighted average, with weights $\{5, 5, 2, 2, 2\}$ respectively \cite{Cao2025CORL}. The glossary of the rule compliance metrics are in Section~\ref{x:exp_protocols} of Appendix.

Similar to the relevant works \cite{wen2024feature}, we also measure how each data selection policy improves EPDMS relative to the \textit{Random} baseline to assess sample efficiency. Specifically, we report the Budget Ratio to Match Random (BRMR); the ratio of the data budget $B$ required by each method to achieve the same EPDMS performance attained by random selection at the same budget. Formally, let $B_k$ denote the number of samples required by selection strategy $k$ to match the EPDMS obtained by random sampling with budget $B$. Then $\text{BRMR}:=B_k / B$. Lower BRMR indicates greater sample efficiency, as it reflects fewer samples needed to achieve the same performance level as random selection.


\begin{table}[h]
\caption{Validation EPDMS (higher is better) and BRMR (lower is better) reported in OpenScene (Section A) and Navtrain (Section B) settings. We report the results for all budgets in Section~\ref{x:more_results} of Appendix.}
\label{tab:main_os_navtrain}
\vspace{-2mm}
 \resizebox{1.0\linewidth}{!}{
\begin{tabular}{@{}cccccc@{}}
\toprule
                                                                            &                                  & \multicolumn{2}{c}{A. Openscene}                                         & \multicolumn{2}{c}{B. Navtrain}                                                           \\ \midrule
Budget                                                                      & \multicolumn{1}{c|}{Method}      & EPDMS ($\uparrow$)                              & \multicolumn{1}{c|}{BRMR ($\downarrow$) }           & EPDMS  ($\uparrow$)                               & BRMR  ($\downarrow$)                                           \\ \midrule
                                                                            & \multicolumn{1}{c|}{Random}      & 72.84\textsubscript{±1.14}          & \multicolumn{1}{c|}{1.00}          & 84.66\textsubscript{±0.60}           & 1.00                        \\
                                                                            & \multicolumn{1}{c|}{Uncertainty} & 70.78\textsubscript{±0.59}          & \multicolumn{1}{c|}{14.58}         & 84.50\textsubscript{±0.48}           & 1.47                        \\
                                                                            & \multicolumn{1}{c|}{Coreset}     & 76.26\textsubscript{±0.48}          & \multicolumn{1}{c|}{0.20}          & 85.29\textsubscript{±0.47}          & 0.53 \\
                                                                            & \multicolumn{1}{c|}{Chameleon}   & 72.97\textsubscript{±1.72}          & \multicolumn{1}{c|}{0.86}          & 84.57\textsubscript{±0.18}          & 1.07                        \\
\multirow{-5}{*}{\begin{tabular}[c]{@{}c@{}}A. 250\\ B. 100\end{tabular}}   & \multicolumn{1}{c|}{\method}     & \textbf{77.38\textsubscript{±1.58}} & \multicolumn{1}{c|}{\textbf{0.15}} & \textbf{86.29\textsubscript{±0.43}} & \textbf{0.30}               \\ \midrule
                                                                            & \multicolumn{1}{c|}{Random}      & 75.84\textsubscript{±0.90}           & \multicolumn{1}{c|}{1.00}          & 86.69\textsubscript{±0.20}           & 1.00                                                \\
                                                                            & \multicolumn{1}{c|}{Uncertainty} & 71.12\textsubscript{±0.38}          & \multicolumn{1}{c|}{8.00}          & 86.07\textsubscript{±0.75}          & 2.00                                                \\
                                                                            & \multicolumn{1}{c|}{Coreset}     & 80.46\textsubscript{±0.02}          & \multicolumn{1}{c|}{0.22}          & 87.09\textsubscript{±0.29}          & 0.79                                                \\
                                                                            & \multicolumn{1}{c|}{Chameleon}   & 79.08\textsubscript{±0.74}          & \multicolumn{1}{c|}{0.49}          & 87.04\textsubscript{±0.60}           & 0.82                                                \\
\multirow{-5}{*}{\begin{tabular}[c]{@{}c@{}}A. 1000\\ B. 400\end{tabular}}  & \multicolumn{1}{c|}{\method}     & \textbf{81.68\textsubscript{±0.52}} & \multicolumn{1}{c|}{\textbf{0.18}} & \textbf{88.21\textsubscript{±0.03}} & \textbf{0.38}                                       \\ \midrule
                                                                            & \multicolumn{1}{c|}{Random}      & 80.38\textsubscript{±0.55}          & \multicolumn{1}{c|}{1.00}          & 88.62\textsubscript{±0.22}          & 1.00                                                \\
                                                                            & \multicolumn{1}{c|}{Uncertainty} & 73.46\textsubscript{±0.19}          & \multicolumn{1}{c|}{2.00}          & 87.75\textsubscript{±0.37}          & 1.36                                                \\
                                                                            & \multicolumn{1}{c|}{Coreset}     & 83.63\textsubscript{±0.36}          & \multicolumn{1}{c|}{0.25}          & 89.30\textsubscript{±0.19}           & 0.58                                                \\
                                                                            & \multicolumn{1}{c|}{Chameleon}   & 82.92\textsubscript{±0.13}          & \multicolumn{1}{c|}{0.39}          & 89.50\textsubscript{±0.20}            & 0.62                                                \\
\multirow{-5}{*}{\begin{tabular}[c]{@{}c@{}}A. 4000\\ B. 1600\end{tabular}} & \multicolumn{1}{c|}{\method }                         & \textbf{84.25\textsubscript{±0.14}} & \multicolumn{1}{c|}{\textbf{0.18}}                      & \textbf{90.18\textsubscript{±0.25}} & \textbf{0.37}                                       \\ \bottomrule
\end{tabular}
}
\vspace{-4mm}
\end{table}

\begin{table*}[h!]
\caption{Breakdown of the nine EPDMS rule-compliance metrics for the base model and the models trained with data selected by various strategies at a single budget, shown for both the OpenScene and Navtrain experiments. } 
\label{tab:epdm_subscores}
\vspace{-2mm}
\resizebox{1.0\linewidth}{!}{
\begin{tabular}{@{}cc|cccc|ccccc|c@{}}
\toprule
\multicolumn{2}{c|}{Setting}                                                               & NC ($\uparrow$)          & DAC ($\uparrow$)             & DDC     ($\uparrow$)          & TLC ($\uparrow$)           & EP ($\uparrow$)              & TTC  ($\uparrow$)                & LK ($\uparrow$)             & HC ($\uparrow$)          & EC   ($\uparrow$)          & EPDMS   ($\uparrow$)           \\ \midrule
A. Openscene                                                                 & Base        & 94.05                               & 83.9                                & 96.28                               & 99.6                                & 85.96                               & 92.95                               & 93.26                               & 98.25                               & 81.88                               & 72.0                                \\ \midrule
\multirow{5}{*}{\begin{tabular}[c]{@{}c@{}}Budget\\ 4000 Clips\end{tabular}} & Random      & 96.32\textsubscript{±0.59}          & 90.53\textsubscript{±0.06}          & 99.06\textsubscript{±0.07}          & 99.79\textsubscript{±0.05}          & 86.36\textsubscript{±0.48}          & 95.66\textsubscript{±0.52}          & 95.68\textsubscript{±0.09}          & 98.30\textsubscript{±0.01}          & 84.46\textsubscript{±0.14}          & 80.38\textsubscript{±0.55}          \\
                                                                             & Uncertainty & 94.67\textsubscript{±0.28}          & 85.11\textsubscript{±0.51}          & 97.15\textsubscript{±0.54}          & 99.71\textsubscript{±0.04}          & 84.26\textsubscript{±0.69}          & 93.72\textsubscript{±0.40}          & 93.26\textsubscript{±0.09}          & 98.28\textsubscript{±0.02}          & 81.34\textsubscript{±1.06}          & 73.46\textsubscript{±0.19}          \\
                                                                             & Coreset     & \textbf{97.11\textsubscript{±0.18}} & 92.93\textsubscript{±0.60}          & 99.44\textsubscript{±0.06}          & \textbf{99.82\textsubscript{±0.02}} & 86.65\textsubscript{±0.55}          & \textbf{96.42\textsubscript{±0.19}} & \textbf{96.66\textsubscript{±0.30}} & 98.16\textsubscript{±0.12}          & 85.10\textsubscript{±0.06}          & 83.63\textsubscript{±0.36}          \\
                                                                             & Chameleon   & 96.76\textsubscript{±0.24}          & 92.32\textsubscript{±0.02}          & 99.51\textsubscript{±0.01}          & 99.77\textsubscript{±0.01}          & 86.98\textsubscript{±0.17}          & 95.91\textsubscript{±0.31}          & 96.49\textsubscript{±0.12}          & \textbf{98.32\textsubscript{±0.01}} & \textbf{85.51\textsubscript{±0.11}} & 82.92\textsubscript{±0.13}          \\
                                                                             & MOSAIC      & 96.97\textsubscript{±0.32}          & \textbf{93.59\textsubscript{±0.11}} & \textbf{99.59\textsubscript{±0.04}} & 99.80\textsubscript{±0.01}          & \textbf{87.14\textsubscript{±0.98}} & 96.18\textsubscript{±0.45}          & 96.62\textsubscript{±0.08}          & 98.28\textsubscript{±0.01}          & 85.06\textsubscript{±0.34}          & \textbf{84.25\textsubscript{±0.14}} \\ \midrule
\multicolumn{1}{l}{B. Navtrain}                                              & Base        & 95.3                                & 95.94                               & 99.09                               & 99.6                                & 88.09                               & 94.55                               & 94.49                               & 98.25                               & 82.39                               & 83.97                               \\ \midrule
\multirow{5}{*}{\begin{tabular}[c]{@{}c@{}}Budget\\ 1600 Clips\end{tabular}} & Random      & 97.17\textsubscript{±0.07}          & 98.19\textsubscript{±0.43}          & 99.42\textsubscript{±0.05}          & 99.69\textsubscript{±0.02}          & 89.36\textsubscript{±0.12}          & 96.50\textsubscript{±0.14}          & 96.45\textsubscript{±0.25}          & \textbf{98.31\textsubscript{±0.03}} & 83.17\textsubscript{±0.76}          & 88.62\textsubscript{±0.22}          \\
                                                                             & Uncertainty & 96.92\textsubscript{±0.38}          & 97.66\textsubscript{±0.08}          & 99.22\textsubscript{±0.10}          & \textbf{99.77\textsubscript{±0.02}} & 89.02\textsubscript{±0.28}          & 96.24\textsubscript{±0.40}          & 96.10\textsubscript{±0.07}          & 98.30\textsubscript{±0.01}          & 82.92\textsubscript{±0.38}          & 87.75\textsubscript{±0.37}          \\
                                                                             & Coreset     & 97.50\textsubscript{±0.10}          & 98.31\textsubscript{±0.34}          & 99.59\textsubscript{±0.03}          & 99.72\textsubscript{±0.05}          & 89.27\textsubscript{±0.21}          & 96.86\textsubscript{±0.07}          & 96.75\textsubscript{±0.22}          & 98.30\textsubscript{±0.03}          & \textbf{83.88\textsubscript{±0.50}} & 89.30\textsubscript{±0.19}          \\
                                                                             & Chameleon   & 97.43\textsubscript{±0.22}          & 98.46\textsubscript{±0.17}          & 99.60\textsubscript{±0.05}          & 99.75\textsubscript{±0.03}          & \textbf{89.60\textsubscript{±0.19}} & 96.83\textsubscript{±0.30}          & 96.89\textsubscript{±0.07}          & 98.30\textsubscript{±0.03}          & 83.87\textsubscript{±0.34}          & 89.50\textsubscript{±0.20}          \\
                                                                             & MOSAIC      & \textbf{98.04\textsubscript{±0.24}} & \textbf{98.61\textsubscript{±0.32}} & \textbf{99.63\textsubscript{±0.06}} & 99.73\textsubscript{±0.02}          & 89.28\textsubscript{±0.19}          & \textbf{97.50\textsubscript{±0.32}} & \textbf{97.07\textsubscript{±0.06}} & 98.28\textsubscript{±0.04}          & 83.70\textsubscript{±0.41}          & \textbf{90.18\textsubscript{±0.25}} \\ \bottomrule
\end{tabular}
}
\end{table*}

\subsection{Main Results: \textbf{\texttt{Openscene}}}

Table~\ref{tab:main_os_navtrain} (Section~A) reports the EPDMS and BRMR scores for different data selection methods. Across all clip budgets $B\in\{250,1000,4000\}$, \textit{MOSAIC} consistently achieves the highest EPDMS, approximately one point higher than the next best method. This demonstrates the superior utility gains of \textit{MOSAIC} under limited data. Moreover, \textit{MOSAIC} requires over \textbf{80\%} fewer samples to match the performance achieved by random selection (i.e., BRMR $<0.2$).


We break down EPDMS into the individual nine metrics in Table~\ref{tab:epdm_subscores}. Section A corresponds to the \openscene\ experiments.
The base model before data collection is particularly limited in DAC and EC, which impact EPDMS. 
\textit{MOSAIC} achieves the largest gain in DAC, nearly 10 points higher than the base, and improves EC and \textit{EP}, while maintaining balanced performance gains across other metrics. In contrast, the other methods yield less gains over the base DAC, and instead improves TTC and EC, that have less effect on the final EPDMS. 
On the other hand, \method\ achieves consistently Top-2 performance across all rule-compliance metrics, while strategically prioritizing DAC, the metric with the greatest room for improvement.
This underscores the importance of incorporating scaling-aware collection into the data selection strategy to optimize $U$ more effectively and achieve better trade-offs across competing metrics.

\subsection{Main Results: \textbf{\texttt{Navtrain}}}
Compared to \openscene, \navtrain\ is a curated dataset emphasizing non-trivial driving scenarios such as dense traffic and complex maneuvers. Table~\ref{tab:main_os_navtrain} (Section B) summarizes the EPDMS and BRMR results. Here, \textit{MOSAIC} consistently delivers the strongest performance, achieving up to 1.1 points higher EPDMS than the next best method across all budgets. It also attains the lowest BRMR values ($<0.4$), corresponding to a 60–70\% reduction in the number of samples needed to match the performance of \textit{Random}. We conclude that \textit{MOSAIC} remains highly effective even on the more challenging, curated \navtrain\ split, where each clip already carries substantial learning value.

The section B of Table~\ref{tab:epdm_subscores} reports the breakdown of EPDMS. \textit{MOSAIC} achieves consistent improvements across all metrics, with the largest gains observed in DAC, NC, and LK. Importantly, \method\ understands the trade-off between the metrics, and shifts the collection effort from saturated, less impactful metrics toward those that require more improvement.
Overall, \method\ provides a more balanced and sustained improvement profile, suggesting that the scaling-aware allocation identifies data with broader generalization benefits. Ultimately, this allows \method\ to achieve a higher EPDMS than the baselines under the same clip budget.



\subsection{Ablating the effectiveness of \method}

\begin{figure}[h]
    \centering
    \includegraphics[width=0.94\linewidth]{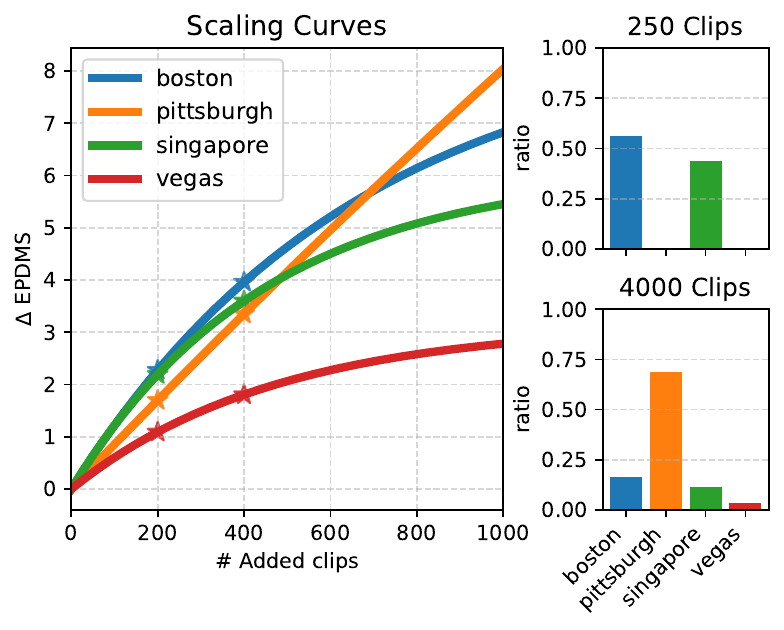}
    \vspace{-4mm}
    \caption{(Left) Performance scalings of different clusters, obtained by fitting the estimator in Equation~\ref{eq:h_i2z} on 2 pilot runs, denoted by $\star$. (Right) Geolocation distributions at different budgets as a result of scaling-aware iterative selection.}
    \label{fig:openscene_map_scalings}
    \vspace{-4mm}
\end{figure}

\paragraph{Dynamics of scaling-aware data selection.} 
In the \openscene~experiments, $\poolset$ is partitioned based on geolocation into four domains corresponding to Las Vegas, Boston, Singapore, and Pittsburgh. Figure~\ref{fig:openscene_map_scalings} illustrates the fitted scaling curves for each city, where $\star$ markers denote the pilot-run results used to estimate the parameters of the scaling curves in Equation~\ref{eq:h_i2z}. 
We note that different domains scale at different rates depending on the how many clips are added. Specifically,  
data collected from Boston and Singapore yield the largest initial performance gains in the low-data regime ($<500$ clips), while Pittsburgh maintains steadier improvements and eventually supercedes all other domains at high data budgets. In contrast, the Las Vegas cluster provides the smallest gains and saturates early. These heterogeneous scaling behaviors are later exploited by the scaling-aware selection policy of \method\ to maximize the performance gain, under varying data budgets.

\begin{figure*}[h]
    \centering
    \includegraphics[width=1.0\linewidth]{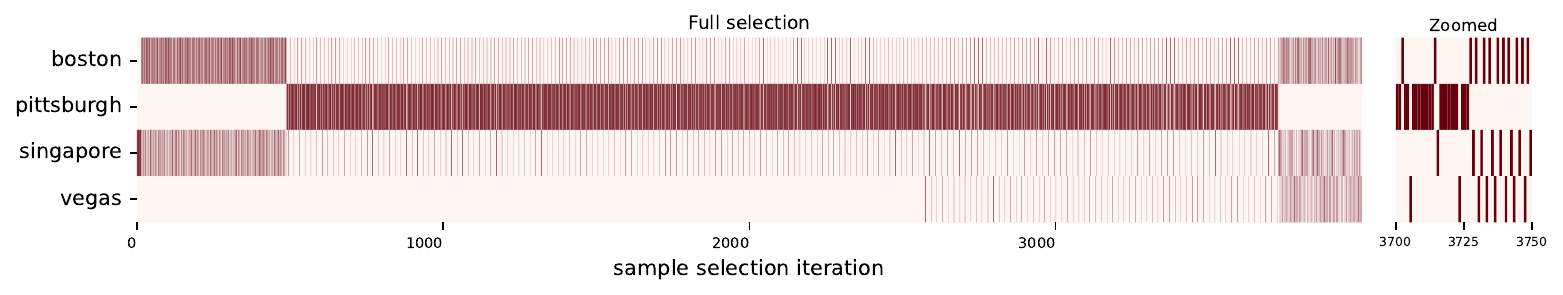}
    \vspace{-7mm}
    \caption{Visualization of the scaling-aware iterative data selection process. The x-axis denotes the sample selection iterations, and the y-axis lists the cluster names. Each vertical bar indicates from which cluster the next sample is mined at a given iteration, based on the estimated cluster-wise scaling fits. The left panel shows the complete selection process up to 4,000 clips, and the right panel zooms into iterations 3,700–3,750 for clarity. }
    \label{fig:sa_selection}
    \vspace{-5mm}
\end{figure*}

Figure \ref{fig:sa_selection} shows how these fitted scaling laws influence the order in which samples are added to the training set. 
The y-axis lists data clusters, i.e. the city names, and the x-axis denotes the iteration index. Each bar indicates from which cluster the next sample is collected from at any iteration. During the early stages, only Boston and Singapore are actively mined, while Las Vegas and Pittsburgh are largely ignored. Figure~\ref{fig:openscene_map_scalings} (top right) confirms this behavior: when the budget is 250, most selected clips originate from Boston and Singapore. As the returns from Boston and Singapore diminish, Pittsburgh’s steadier scaling curve makes it increasingly favorable between indices 500 to 3700. Figure~\ref{fig:openscene_map_scalings} (bottom right) shows that at 4000 clips, the selected set is dominated by Pittsburgh samples. After around 3700 collection rounds, the Pittsburgh data domain is exhausted. Beyond approximately 2500 sample selections, the scaling curves of Boston, Singapore, and Pittsburgh approach saturation, causing their marginal gains to diminish. As a result, the expected improvement from the initial Las Vegas samples becomes comparable to those of the other regions, leading \textit{\method} to mine from Vegas.


\begin{figure}[h]
\centering
\includegraphics[width=0.9\linewidth]{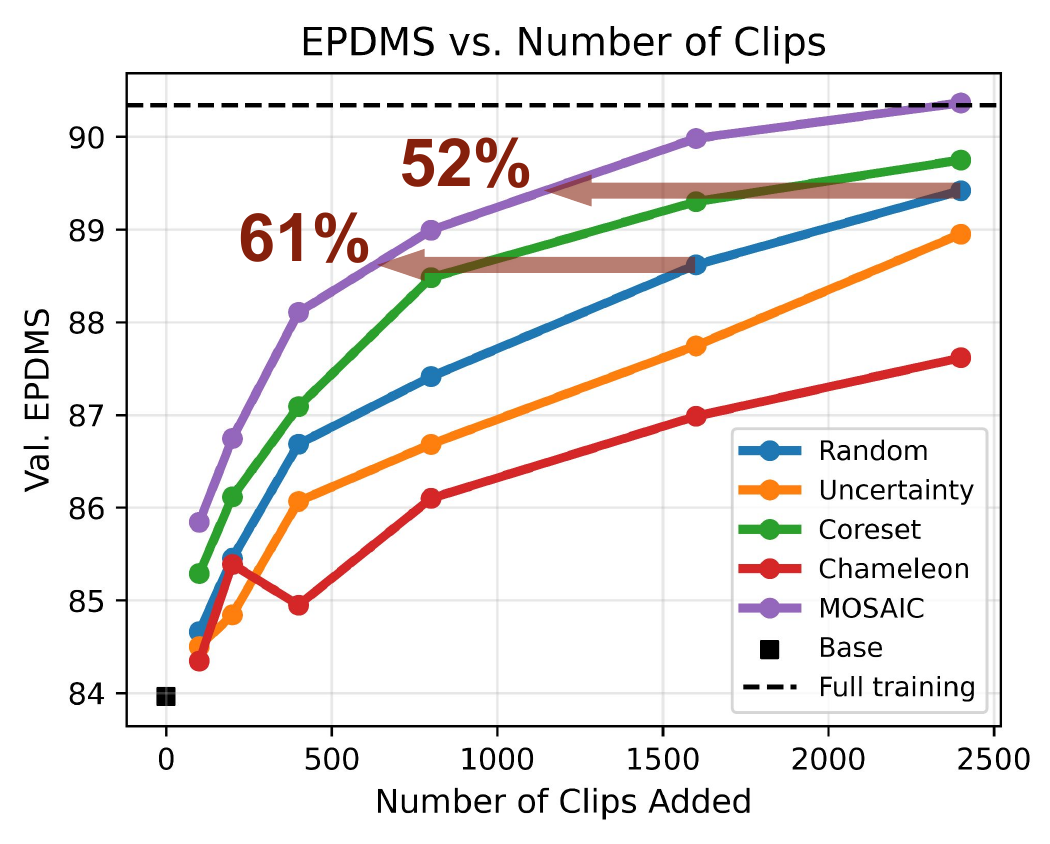}
\vspace{-4mm}
\caption{Validation EPDMS for various budgets obtained by different strategies when \texttt{Navtrain} data pool is clustered using clip captions. \textit{MOSAIC} requires 61\% and 52\% fewer clips than \textit{Random} selection to match its performance at the 1,600- and 2,400-clip budgets, respectively.}
\label{fig:caption_cluster_val_epdms}
\vspace{-6mm}
\end{figure}

\vspace{-3.5mm}
\paragraph{MOSAIC is optimal under different clustering mechanisms.}
Instead of partitioning the data pool into domains separated by geolocation, we cluster the \navtrain\ data pool using captions generated for each clip. We use the Qwen-2.5-VL-32B-Instruct model~\cite{Qwen2.5-VL} to generate captions for all clips in $\mathcal{D}_{pool}$. We form six clusters that capture distinct driving and scene contexts using the TF-IDF feature vectors of the generated captions. The top uni-grams and bi-grams characterizing each cluster are provided in Table~\ref{tab:caption_clusters}. 

\begin{table}[t]
\centering
\caption{Top uni-grams and bi-grams of different clusters. }
\label{tab:caption_clusters}
\vspace{-2mm}
\resizebox{0.9\linewidth}{!}{
\begin{tabular}{@{}cl@{}}
\toprule
   &  Uni-grams \& Bi-grams                       \\ \midrule
Cluster 1 & calm, day, street, trees, signs, yellow         \\
Cluster 2 & signals, crossing, crosswalks, pedestrians      \\
Cluster 3 & highway, vehicles, busy urban, palm trees       \\
Cluster 4 & building, area, large, paved, parking           \\
Cluster 5 & city street, major city, moderate               \\
Cluster 6 & precipitation, potential rain, overcast, cloudy \\ \bottomrule
\end{tabular}
}
\vspace{-3mm}
\end{table}

Figure~\ref{fig:caption_cluster_val_epdms} reports the validation EPDMS as a function of the data selection budget for all strategies. We also indicate the base performance and the full-training performance, corresponding to the model trained by including all 4,141 clips in $\poolset$ for training. (We provide full table with subscores in Section~\ref{x:more_results} of Appendix)
\textit{MOSAIC} consistently outperforms all baselines, including \textit{Chameleon} (i.e., the other data mixture optimization method); requires 61\% and 52\% fewer samples than random selection to match its performance at the highest budgets of 1,600 and 2,400 clips, respectively. Moreover, \textit{MOSAIC} reaches the full performance of training with all data samples using only 2,400 clips, i.e., 42\% fewer samples. 
Interestingly, \textit{Chameleon} degrades under caption-based clustering, despite being the strongest baseline in the previous setting. This indicates that its kernel ridge weighting is highly sensitive to the structure of the clustered domains. 
Also, since the clustering choice only affects \textit{Chameleon} and \textit{MOSAIC}, the other strategies have the same performance as before.


\vspace{-3mm}
\paragraph{Combining clustering and ranking yields the best data selection policy.}
We ablate the effects of both clustering and ranking components by individually disabling them. First, we use a ``\textit{w/o Clustering}'' variant where we simply rank $\mathcal{D}_{pool}$ by the EPDMS importance scores $\importance(x)$ and greedily add samples with the lowest scores until reaching the collection budget. Second, we use a ``\textit{w/o Ranking}'' variant where we disable the ranking step, while retaining clustering and the scaling-aware estimation of how many samples to collect. Here, we simply sample data points from each domain randomly to satisfy the budget. Moreover, the scaling laws are also estimated on unranked domains.

\begin{figure}[h]
\centering
\includegraphics[width=0.8\linewidth]{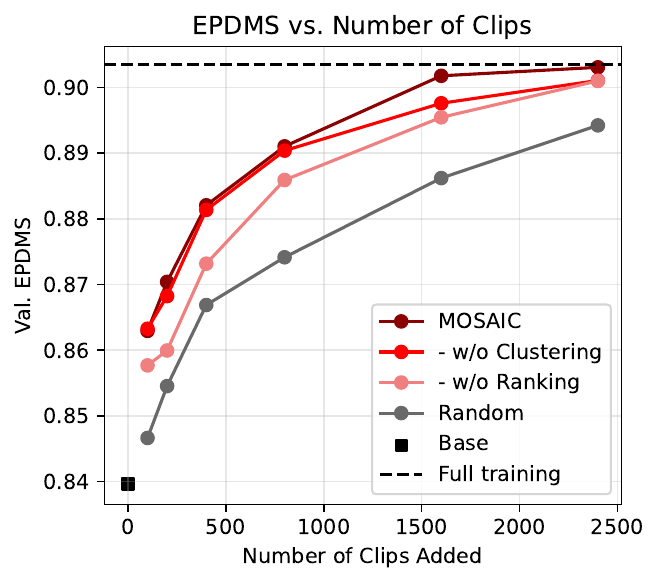}
\vspace{-3mm}
\caption{Analyzing the contribution of different components of the \method\ framework.}
\label{fig:ablation_method_components}
\vspace{-7mm}
\end{figure}

Figure \ref{fig:ablation_method_components} visualizes these baselines to show that performance improvements in the low-data regime (up to a collection budget of 800 clips) can largely be attributed to ranking. The \textit{\method} and \textit{w/o Clustering} variants achieve competitive EPDMS in this region. However, in the higher data regime, merely adding clips with low EPDMS scores becomes less effective, as the performance of \textit{w/o Clustering} begins to lag behind \textit{\method}. Finally, we note that both of these disabled variants still outperform random collection by a large margin. 

\section{Limitations} 
We note two key limitations of \method. First, the relaxation from Equation~\ref{eq:mixture_optim} to Equation~\ref{eq:separable_approx} assumes that each cluster’s contribution is well captured by its own scaling curve $\Delta U_i(n)$, with limited cross-cluster interactions. Consequently, if the clustering fails to produce well-separated groups, this assumption may be violated, leading \method\ to suboptimal allocation. 

Second, \method\ relies on pilot runs to estimate cluster-specific scaling curves, which introduces additional computational cost. However, as shown in Section~\ref{x:scaling_fits} of the Appendix, accurate scaling fits can be obtained efficiently using small pilot subsets or through continual training. Thus, despite the initial overhead for the pilot runs to obtain cluster scalings, \method\ ultimately requires less total compute to achieve superior performance in the large-data regime.


\section{Conclusion}

We introduce \method, a scaling-aware data selection framework that jointly leverages clustering, ranking, and scaling-law modeling to maximize the performance of a model defined by multiple competing metrics, under a limited data budget. We apply \method\ to E2E AD, where a planner model uses a diverse data pool to optimize a utility function that aggregates competing rule compliance metrics. Empirically, \method\ consistently outperforms existing data selection and mixture baselines on both the \openscene\ and \navtrain\ datasets by achieving substantial gains in EPDMS and sample efficiency. Ablation studies further highlight the framework’s mechanisms, analyze the necessity of the individual components components, and demonstrate robustness to clustering choices as long as semantic consistency is maintained. Overall, \method\ offers a general and principled blueprint for identifying influential data in large-scale, heterogeneous learning systems.

{
    \small
    \bibliographystyle{ieeenat_fullname}
    \bibliography{main}
}

\clearpage
\setcounter{page}{1}
\maketitlesupplementary

\section{Experiment Protocols}
\label{x:exp_protocols}

\subsection{Dataset and Virtual Clip Creation}
\label{x:exp_setup:dataset}

We conduct experiments using the \texttt{Navtrain}~\cite{dauner2024navsim} and \texttt{trainval} splits of \texttt{OpenScene}~\cite{contributors2023openscene} as the combined training and pool datasets.
\texttt{OpenScene} is a redistribution of the NuPlan dataset~\cite{nuplan}, subsampled to 2 Hz, and contains approximately 120 hours of driving data with dense annotations.
The \texttt{Navtrain} split is curated within the NAVSIM framework~\cite{dauner2024navsim} by filtering out trivial driving scenarios from the \texttt{trainval} split of \texttt{OpenScene}.
In both experiments, evaluation is performed on \texttt{navtest}~\cite{dauner2024navsim}, a validation set curated analogously from the \texttt{test} split of \texttt{OpenScene}.

The \texttt{Navtrain} and \texttt{trainval} splits consist of 1,192 and 1,250 individual driving sessions, respectively, with durations ranging from 30 seconds to 50 minutes. In addition to this large temporal variation, the total number of available driving sessions remains limited, and treating individual frames as independent samples to would be both unrealistic and inconsistent with the temporal structure of driving data. Hence, to have more samples to work with and to align our data handling with common industry practice \cite{waymo_open_motion_dataset} we segment each driving log into fixed-length \textit{virtual clips} of 10 seconds (corresponding to 20 frames at 2 Hz). Below, we describe how we create virtual clips are created.

\begin{table}[h]
\centering
\caption{Train-pool clip counts for OpenScene and Navtrain}
\label{x:table:train_pool}
\begin{tabular}{@{}ccc@{}}
\toprule
          & Train & Pool  \\ \midrule
Openscene & 1000  & 31539 \\
Navtrain  & 460   & 4141  \\ \bottomrule
\end{tabular}
\end{table}

We segment each driving log into fixed-length \textit{virtual clips} of 10 seconds. Given the dataset’s sampling rate of 2 Hz, each virtual clip contains 20 frames. For each log, non-overlapping clips are extracted sequentially from the start of the log, and any remaining portion shorter than 10 seconds is discarded. For example, a 23-second log yields two clips covering [0–10) s and [10–20) s, while the final 3 seconds are omitted. Following this procedure, the \texttt{Navtrain} split yields a total of 4,601 virtual clips after discarding 11,268 out of 103,288 frames (10.9\%). For \texttt{OpenScene}, we obtain 32,539 virtual clips, with 12,086 out of 662,866 frames (1.8\%) omitted due to incomplete segments. The train-pool clip counts are summarized in Table~\ref{x:table:train_pool}

\subsection{Training details}

As already mentioned in the main body of the paper, we use the Hydra-MDP model \cite{li2024hydra} that won the NAVSIM benchmark in 2024 \cite{dauner2024navsim} by a significant margin. In addition to the imitation trajectory loss, the model distills the rule-compliance scores of each trajectory, obtained with prior simulations. We initialize the model's encoder as pretrained VoVNetV2-99 backbone~\cite{lee2020centermask, park2021pseudo}. In accordance with the training recipe of provided in the paper \cite{li2024hydra}, we use Adam optimizer \cite{KingmaBa2015} without any weight decay and keep the learning rate fixed throughout the training. We set the per-GPU batch size to 20. 

In the \texttt{Navtrain} experiments, all runs are conducted using 8$\times$A100 GPUs with a learning rate of $1\text{e}{-4}$. Each experiment is repeated with three random seeds (0, 2025, 424242), and the reported results are averaged over these runs, with the standard deviation shown as a subscript. The base experiment and budgets up to 800 clips are trained for 60 epochs, while the 1,600- and 2,400-clip settings are trained for 50 and 45 epochs, respectively, to reduce compute cost.

For the \texttt{OpenScene} experiments, the rule-compliance distillation losses are disabled due to their high computational overhead needed to run intensive simulations to calculate those scores. All runs use 16$\times$A100 GPUs with a learning rate of $2\text{e}{-4}$ and a fixed training length of 40 epochs. Experiments with 2,000, 4,000, and 8,000 clips are repeated with two seeds (0, 2025), while smaller-budget runs use three seeds (0, 2025, 424242) to ensure stability.

\subsection{Details of the Baselines}

\paragraph{Random.} For each budget $B$, Random selection is constructed from a single randomized ordering of the pool. Specifically, we shuffle all clips once using a fixed seed (seed = 42) and define the selected set for budget $B$ as the first $B$ clips in this ordering. This ensures that selections for larger budgets are strict supersets of those for smaller budgets.

\paragraph{Uncertainty~\cite{joshi2009multi}.} We score each pool clip by the entropy of its model-predicted trajectory logits. Let $z_i$ denote the (pre-softmax) logits for sample $x_i$, and let $p_i = \text{softmax}(z_i)$ be the corresponding probability distribution over candidate trajectories. The uncertainty score is taken as the Shannon entropy
$H_i = -\sum_k p_{i,k} \log p_{i,k}$. The uncertainty score is calculated for each frame in the clip, and we simply take average of the frame uncertainty scores to aggregate it at the clip level. Clips with higher entropy correspond to more ambiguous or uncertain model predictions and are therefore preferred. To construct a budget-$B$ selection, we compute $H_i$ for every pool item once, rank all items by entropy in descending order, and pick the top $B$. We share the procedure in Algorithm~\ref{alg:uncertainty_entropy}.

\begin{algorithm}[h]
\caption{Entropy-Based Uncertainty Selection}
\label{alg:uncertainty_entropy}
\begin{algorithmic}[1]
\Require Pool samples $\{x_i\}$, model $f(\cdot)$, budget $B$
\Ensure Selected set $S$ of size $B$
\State Initialize $S = \emptyset$
\For{each sample $x_i$ in the pool}
    \State $z_i = f(x_i)$ \Comment{trajectory logits}
    \State $p_i = \mathrm{softmax}(z_i)$
    \State $H_i = - \sum_k p_{i,k} \log p_{i,k}$ \Comment{entropy score}
\EndFor
\State Rank all pool samples by $H_i$ in descending order
\State $S \gets$ top-$B$ samples under this ranking
\State \Return $S$
\end{algorithmic}
\end{algorithm}

\paragraph{Coreset~\cite{sener2017active}.} We adopt the standard geometric Coreset selection procedure shown in Algorithm~\ref{alg:coreset_standalone}. Starting from an initial set of training indices $s^0$, the algorithm iteratively adds the pool element that is farthest under the chosen distance measure $\Delta(\cdot,\cdot)$ from the current selected set. Specifically, we use Euclidean Distance. At each iteration, Coreset identifies the sample $u \in s^{pool}$ that maximizes the minimum distance to the existing set $s$, and then augments $s$ with $u$. This expansion continues until the total size reaches $B + |s^0|$, yielding the Coreset of size $B$ from the pool.

\begin{algorithm}[h]
\caption{Coreset}
\label{alg:coreset_standalone}
\begin{algorithmic}[1]
\Require train sample indices $s^0$, budget $B$, pool indices $s^{pool}$
\State Initialize $s = s^0$
\Repeat
    \State $u = \arg \max_{i \in s^{pool}} \min_{j \in s} \Delta(x_i, x_j)$
    \State $s = s \cup \{u\}$
\Until{$|s| = B + |s^0|$}
\State \Return $s$
\end{algorithmic}
\end{algorithm}

\paragraph{Chameleon~\cite{xie2025chameleon}.} Chameleon is a domain-mixture framework that relies on embeddings computed from the training domains. First, each cluster is embedded using representations from the base model’s feature space. For each cluster, sample embeddings are averaged which produces one embedding per cluster. A cluster–cluster affinity matrix is then constructed using a kernel function applied to pairs of domain embeddings. Given this affinity matrix, Chameleon applies kernel ridge regression (KRLS) to compute a score $S_i$ for each domain, reflecting how informative or influential that domain is relative to all others. Finally, the mixture weight for domain $i$ is obtained by normalizing these scores with a softmax, $\alpha_i = \mathrm{softmax}(S_i)$, and data are sampled from domains according to these mixture weights. We use the pretraining mode in our experiments, as we re-train the model from scratch for each budget and we set the ridge parameter as $\lambda=1$. The pseudo-code is provided in Algorithm~\ref{alg:chameleon_pretraining}.

\begin{algorithm}[h]
\caption{Chameleon Domain Weighting (Pretraining Mode)}
\label{alg:chameleon_pretraining}
\begin{algorithmic}[1]
\Require Training clusters $\mathcal{D} = \{D_1,\ldots,D_k\}$, ridge parameter $\lambda$, embedding layer $L$, budget $B$
\Ensure Selected set $S$ of size $B$
\State Extract domain embeddings:
\State \hspace{1em} $x_i = \frac{1}{|D_i|} \sum_{a \in D_i} h^{(L)}_\theta(a)$ for each domain $D_i$
\State Construct feature matrix $X = [x_1^\top,\ldots,x_k^\top]$
\State Compute affinity matrix $\Omega_D = X X^\top$
\State Compute KRLS scores $S_\lambda(D_i)$ for each domain $D_i$ using $\Omega_D$
\State Compute domain weights:
\State \hspace{1em} $\alpha_i^{PT} = \frac{\exp(S_\lambda^{-1}(D_i))}{\sum_{j=1}^k \exp(S_\lambda^{-1}(D_j))}$
\State Sample $B$ points from domains according to mixture weights $\{\alpha_i^{PT}\}$
\State \Return $S$
\end{algorithmic}
\end{algorithm}

\section{More Results on the Experiments and Ablations}
\label{x:more_results}

Due to the space constraints in the main body of the paper, we present more results here.

\begin{figure}[h]
\centering
\includegraphics[width=0.8\linewidth]{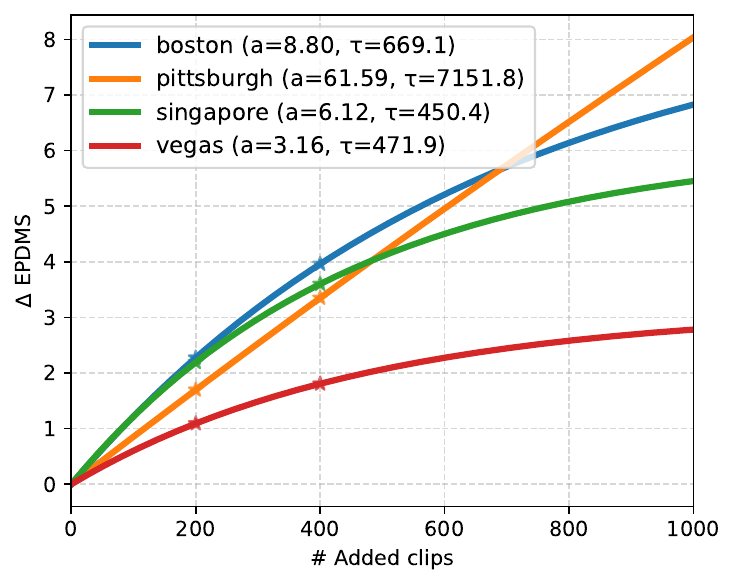}
\vspace{-3mm}
\caption{Performance scalings of different cities for the OpenScene experiment.}
\label{x:fig:scaling_openscene}
\end{figure}

\paragraph{Experiments on \textbf{\openscene}.} The full validation EPDMS and BRMR results for the Openscene experiments can be found in Table~\ref{x:tab:main_openscene}. The breakdown of the validation EPDMS subscores are shared in Table~\ref{x:tab:breakdown_openscene}. The scaling curves obtained from different cities are shared in Figure~\ref{x:fig:scaling_openscene}.

\begin{figure}[h]
\centering
\includegraphics[width=0.8\linewidth]{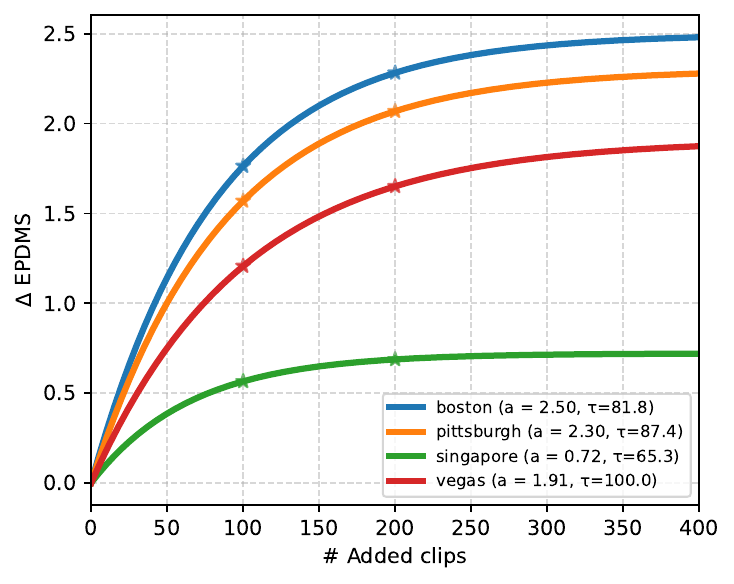}
\vspace{-3mm}
\caption{Performance scalings of different cities for the main Navtrain experiment.}
\label{x:fig:scaling_navtrain}
\end{figure}

\begin{figure}[h!]
    \centering

    \begin{subfigure}[t]{0.49\linewidth}
        \centering
        \includegraphics[width=\linewidth]{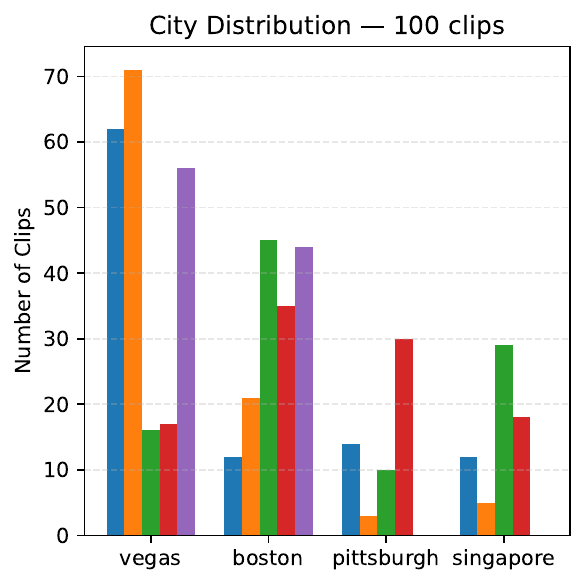}
        \caption{100 clips}
    \end{subfigure}
    \hfill
    \begin{subfigure}[t]{0.49\linewidth}
        \centering
        \includegraphics[width=\linewidth]{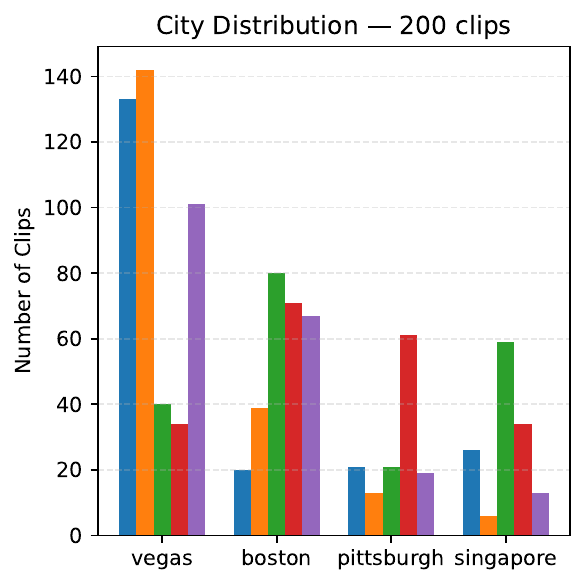}
        \caption{200 clips}
    \end{subfigure}

    \vspace{0.6em}

    \begin{subfigure}[t]{0.49\linewidth}
        \centering
        \includegraphics[width=\linewidth]{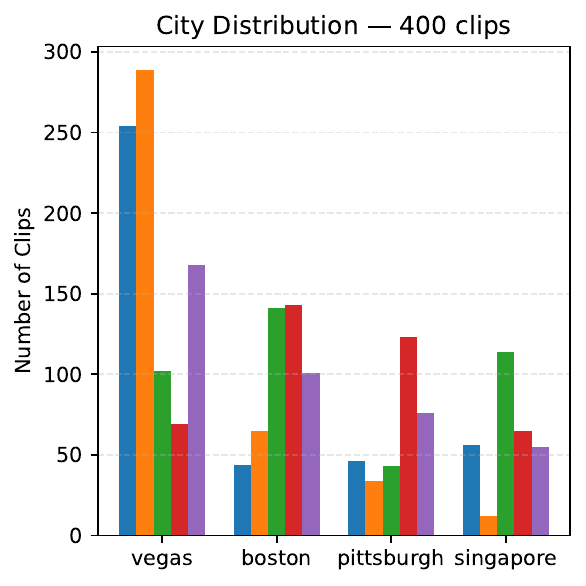}
        \caption{400 clips}
    \end{subfigure}
    \hfill
    \begin{subfigure}[t]{0.49\linewidth}
        \centering
        \includegraphics[width=\linewidth]{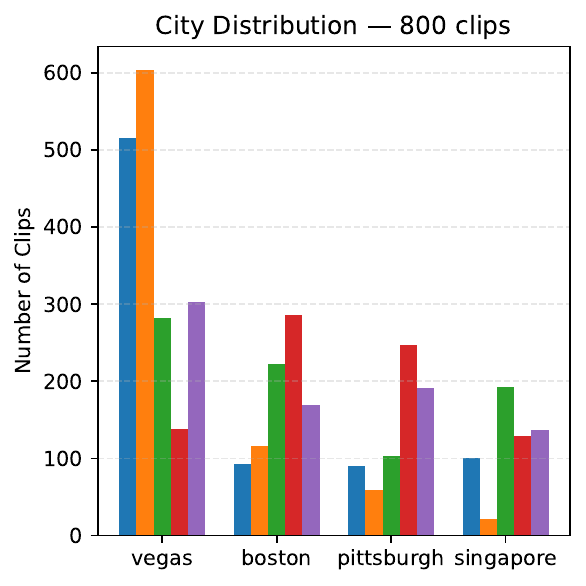}
        \caption{800 clips}
    \end{subfigure}

    \vspace{0.6em}

    \begin{subfigure}[t]{0.49\linewidth}
        \centering
        \includegraphics[width=\linewidth]{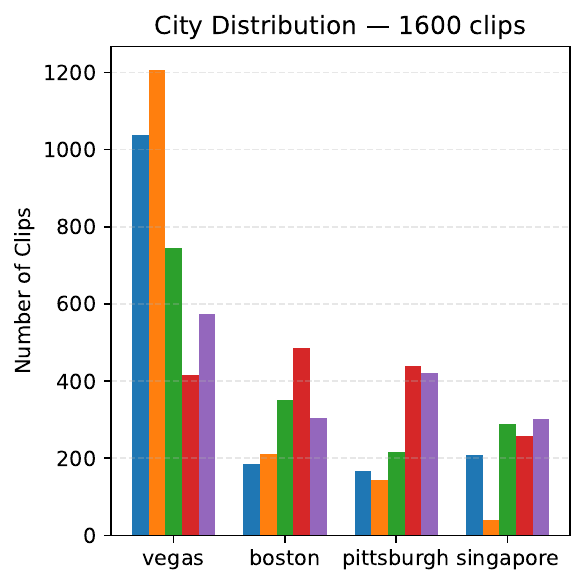}
        \caption{1600 clips}
    \end{subfigure}
    \hfill
    \begin{subfigure}[t]{0.49\linewidth}
        \centering
        \includegraphics[width=\linewidth]{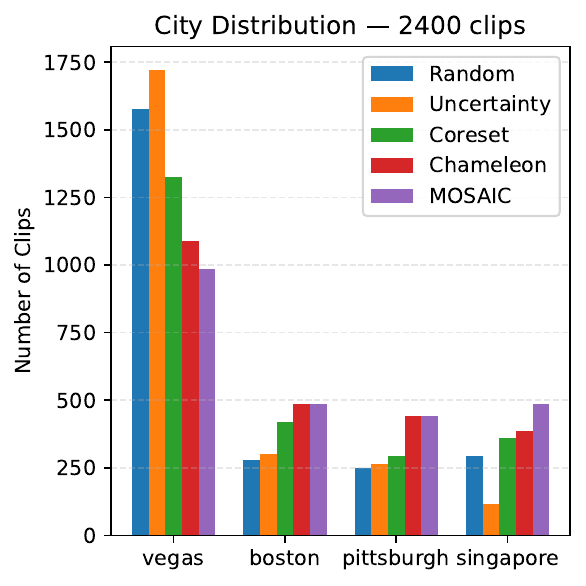}
        \caption{2400 clips}
    \end{subfigure}

    \caption{Overall caption describing all six subfigures.}
    \label{x:fig:navtrain_geo_dists}
\end{figure}

\paragraph{Experiments on \textbf{\navtrain}.} The full validation EPDMS and BRMR results for the Navtrain experiments can be found in Table~\ref{x:tab:main_navtrain}. The breakdown of the validation EPDMS subscores are shared in Table~\ref{x:tab:breakdown_navtrain}. We also provide the city distributions induced by different method at various budgets in Figure~\ref{x:fig:navtrain_geo_dists}. The scaling curves obtained from different cities are shared in Figure~\ref{x:fig:scaling_navtrain}. The scaling curves obtained from different cities are shared in Figure~\ref{x:fig:scaling_navtrain}.

\paragraph{Ablation with Caption-based Clustering.} To generate the clip captions, we used the Qwen-2.5VL-32B-Instruct model with the following caption: ``\textit{This is a 10 second long video of your student driving. The clip might include discontinuities, sudden changes in the driving environment. \
                                            Describe the driving environment that your student is driving through and your student's driving actions. \
                                            Please describe the driving condition including the location, weather, road users, and their motions. During your description, there are several things to keep in mind. \
                                            1. Please pay attention only to the objects on the driving roads and ignore the background. \
                                            2. Ignore the brands of the vehicles. \
                                            3. Describe it if objects are partially occluded by others, or are in areas with different brightness such as under shades. \
                                            Please provide a concise description in one paragraph with less than 150 words. \
                                            Do not mention anything that you are certain does not exist! No statements about uncertain objects or events (no 'maybe' or 'might' or 'possibly'). \
                                            All responses must be in English only!}''

On the generated clip captions, we extract TF–IDF features using the top 1,024 unigrams and bigrams after removing common English stop words. We then perform clustering in this TF–IDF space, forming six clusters. The dominant scene characteristics of each cluster are determined by their highest-weight unigrams and bigrams, as summarized in Table~\ref{tab:caption_clusters}. We additionally conduct a qualitative assessment of the resulting groups and confirm that the clusters are coherent and semantically meaningful.

In fact, we have first attempted using "sentence-transformers/all-mpnet-base-v2" model downloaded from Huggingface to obtain caption embeddings using a pretrained transformer. However, when we clustered the data in this embedding space, qualitative inspection revealed that the resulting groups lacked coherent driving characteristics. Hence, we experimented with clustering on the TF-IDF features which produced much more coherent clusters with directly interpretable feature space.


\begin{figure}[h!]
    \centering

    \begin{subfigure}[t]{0.49\linewidth}
        \centering
        \includegraphics[width=\linewidth]{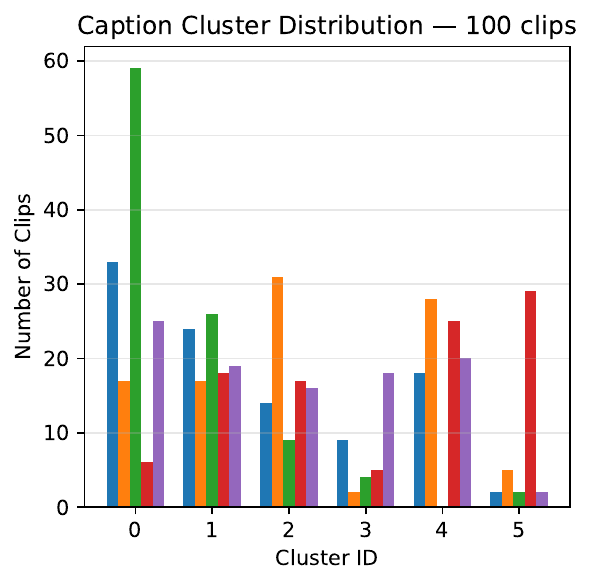}
        \caption{100 clips}
    \end{subfigure}
    \hfill
    \begin{subfigure}[t]{0.49\linewidth}
        \centering
        \includegraphics[width=\linewidth]{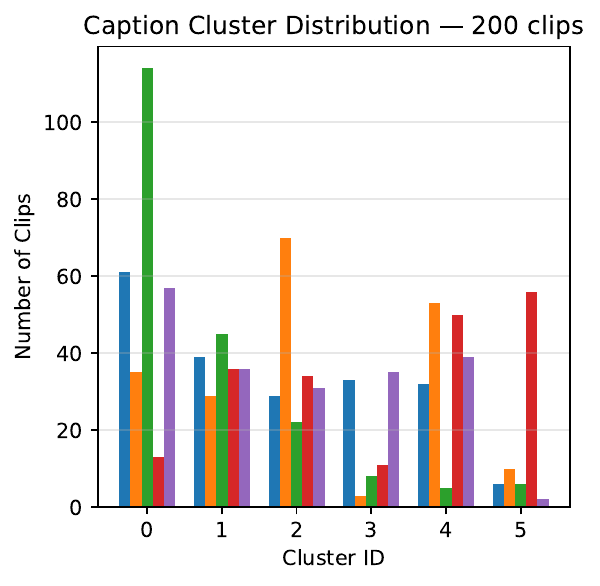}
        \caption{200 clips}
    \end{subfigure}

    \vspace{0.6em}

    \begin{subfigure}[t]{0.49\linewidth}
        \centering
        \includegraphics[width=\linewidth]{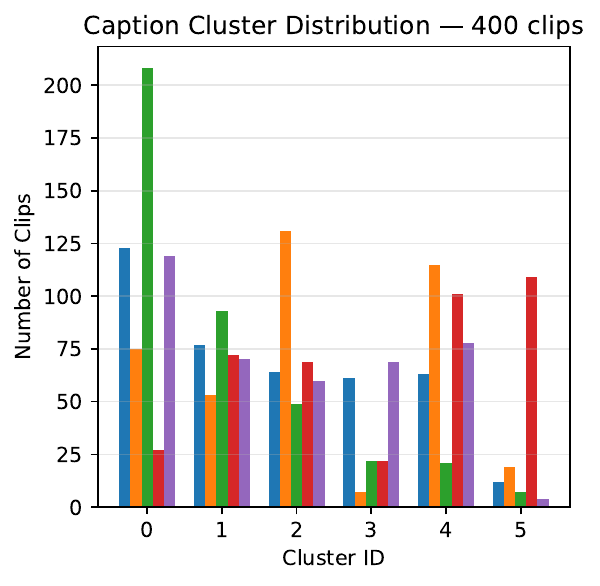}
        \caption{400 clips}
    \end{subfigure}
    \hfill
    \begin{subfigure}[t]{0.49\linewidth}
        \centering
        \includegraphics[width=\linewidth]{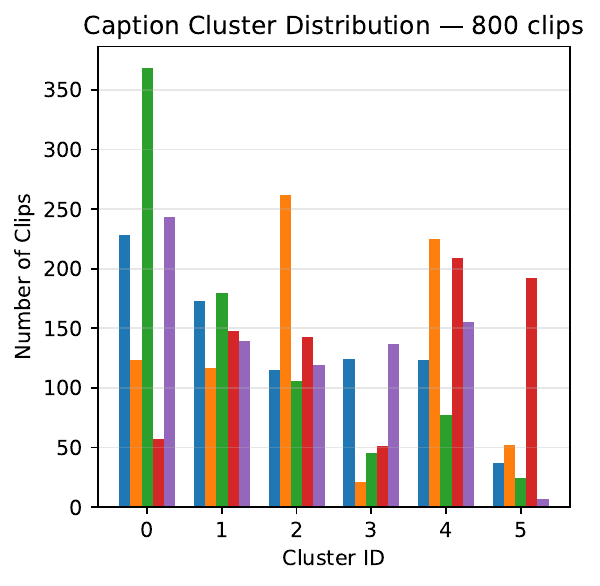}
        \caption{800 clips}
    \end{subfigure}

    \vspace{0.6em}

    \begin{subfigure}[t]{0.49\linewidth}
        \centering
        \includegraphics[width=\linewidth]{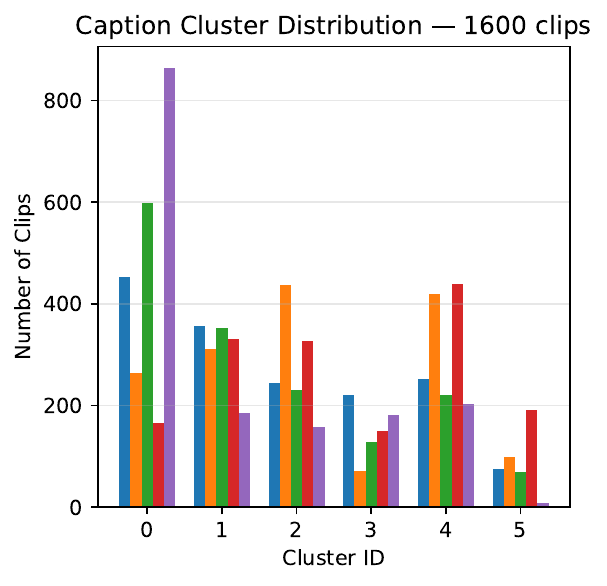}
        \caption{1600 clips}
    \end{subfigure}
    \hfill
    \begin{subfigure}[t]{0.49\linewidth}
        \centering
        \includegraphics[width=\linewidth]{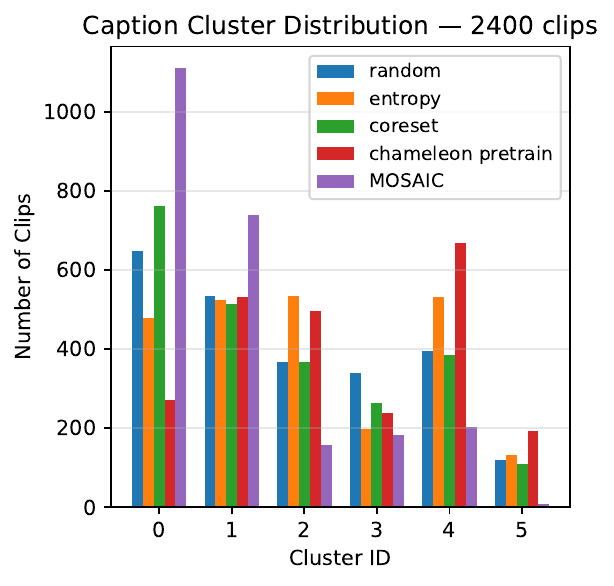}
        \caption{2400 clips}
    \end{subfigure}

    \caption{Overall caption describing all six subfigures.}
    \label{x:fig:navtrain_caption_dists}
\end{figure}

\section{Details on the Scaling Fits and Compute Budget.}
\label{x:scaling_fits}

MOSAIC requires an upfront compute investment to estimate cluster-specific scaling curves via pilot runs. To keep this cost tractable, we avoid full training from-scratch during the pilot experiments. Instead, we adopt a continual-training approach: we resume training from the base model’s final epoch checkpoint and fine-tune on the combined dataset for a small number of epochs. For the OpenScene experiments, we train for 5 epochs after mining 200 and 400 clips from each cluster. For the Navtrain experiments, we train for 10 epochs after mining 100 and 200 clips in the two pilot runs. This procedure provides accurate scaling estimates while maintaining a manageable computational overhead.

\begin{figure}[h]
\centering
\includegraphics[width=0.9\linewidth]{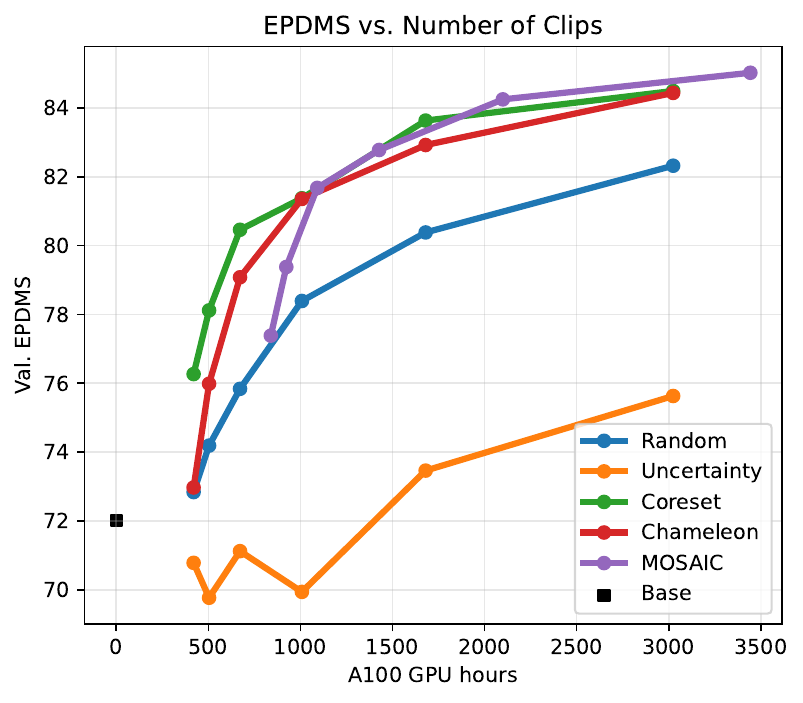}
\vspace{-3mm}
\caption{Validation EPDMS vs. Compute Spent (GPU hours) for OpenScene experiments.}
\label{fig:compute_cost}
\end{figure}

For the OpenScene experiments, we share the results with respect to the compute spent for each method. In particular, we provide the validation EPDMS vs. A100 GPU hours. The results are shared in Figure~\ref{fig:compute_cost}. As can be seen while MOSAIC is not the strongest method at small compute budgets, its initial scaling overhead amortizes over time, and at large budgets, the investment in scaling pays off, making MOSAIC the top-performing approach. More concretely, at the highest compute budget: MOSAIC reaches the top-baseline(Coreset in this setting) performance with 16\% less compute, corresponding to ~490 GPU hours saved; Compared to Random selection, MOSAIC requires 57\% less compute, saving ~1700 GPU hours to attain the same EPDMS. These results demonstrate that although MOSAIC pays an upfront cost for pilot scaling runs, the compute investment is recovered once we move into the large-budget regime.

\section{Ranking with Alternative Cheap Signals}

Since ranking is one of the key components of our framework, we also investigate cheaper alternatives to the EPDMS-based ranking signal to reduce the reliance on dense annotations such as bounding boxes. Specifically, we experiment with ranking clips according to (i) the trajectory imitation loss, (ii) the norm of the gradient vector induced by this loss, and (iii) the sensitivity of the model’s output to gradient perturbations. 

\begin{figure}[h]
\centering
\includegraphics[width=0.8\linewidth]{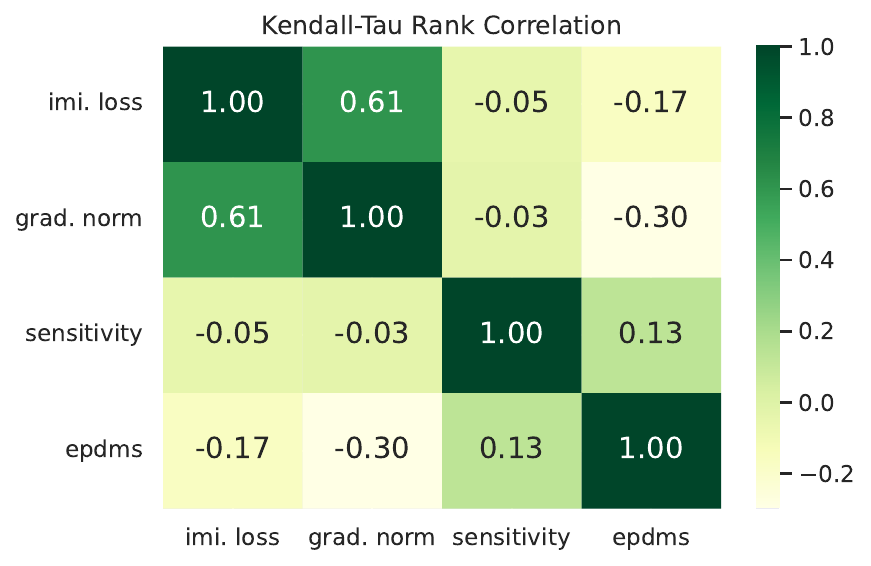}
\vspace{-3mm}
\caption{Kendall-Tau correlation coefficients between EPDMS and cheap signals based rankings.}
\label{fig:ablation_cheap_corr}
\end{figure}

Instead of retraining the model with clips selected using the alternative signals and reporting the validation EPDMS, we measure the Kendall–Tau correlation coefficient between the rankings produced by each alternative signal and those produced by the EPDMS-based ranking. The results, shown in Figure~\ref{fig:ablation_cheap_corr}, indicate that none of the inexpensive alternatives yield a ranking that correlates strongly with EPDMS.

\section{Approximation for Linear Separability and Error Analysis:} Here, we formally express the performance improvement obtained from a data mixture $\Delta U(n_1, \cdots, n_M)$ as follows:
\vspace{-1mm}
\begin{equation*}
       \sum_{i=1}^M \Delta U_i(n_i) + \sum_{i\neq j}\Delta U_{ij}(n_i, n_j) + \text{ H.O.T.}
\end{equation*}
\vspace{-2mm}

Here, the pairwise cross-cluster interaction term $\Delta U_{ij}(n_i, n_j)$ is defined as $\Delta U_{ij}=U_{ij} - U_i - U_j + U_0,$ where we use a lightweight notation for clarity:
$U_{ij} \!=\! U(\trainset \cup \set{D}_{sel}^i \cup \set{D}_{sel}^j)$,
$U_i \!=\! U(\trainset \cup \set{D}_{sel}^i)$,
and $U_0 \!=\! U(\trainset)$,
with $U(\cdot) \equiv U(\{\set{G}_r(\cdot)\}_{r=1}^R)$. In Equation~\ref{eq:separable_approx}, we retain only the first-order terms $\{\Delta U_i\}_{i=1}^M$ and omit interaction and higher-order terms. Importantly, we do not assume strict linear separability. Rather, we assume that first-order cluster-wise scaling captures the dominant variation in performance, while interaction terms contribute residual approximation error. 

To quantify the magnitude of the approximation error, we compare the \textit{estimated} EPDMS calculated by summing cluster-wise scaling fits against the \textit{actual} EPDMS obtained with the MOSAIC data mixtures. As shown in Table~\ref{tab:rebuttal}, the approximation overestimates performance by a modest margin (up to 1 EPMS), indicating that interaction terms are present but negligible in this setting.

\vspace{-2mm}
\begin{table}[h]
\caption{Actual vs. estimated EPDMS (Navtrain, geolocation)}
\label{tab:rebuttal}
\centering
\begin{tabular}{c|lccccc}
\# Clips  & \multicolumn{1}{c}{100} & 200  & 400  & 800  & 1600 & 2400 \\ \hline
\textit{Actual} & 86.3                    & 87.1 & 88.2 & 89.1 & 90.2 & 90.3 \\
\textit{Estimated} & 86.2                    & 87.6 & 89.3 & 90.6 & 91.1 & 91.3
\end{tabular}
\end{table}
\vspace{-2mm}
We also note that the discrepancy between the \textit{Actual} and \textit{Estimated} are accumulation of two factors: i) the cross-cluster interactions, ii) extrapolation errors of the scaling fits. Hence, Table~\ref{tab:rebuttal} should be interpreted as an upper bound on interaction effects rather than a pure estimate thereof.

Also, as a contrasting example, if clusters were formed randomly and lacked semantic coherence, cross-cluster interactions would likely be large, and the approximation would break down. In such pathological settings, explicitly modeling interaction terms would be necessary for optimal data selection.

\clearpage

\begin{table}[]
\centering
\caption{Openscene validation EPDM and BRMR results.}
\label{x:tab:main_openscene}
\resizebox{0.8\linewidth}{!}{
\begin{tabular}{@{}cccc@{}}
\toprule
Budget                & Method      &  EPDMS           & SRR  \\ \midrule
\multirow{5}{*}{250}  & Random      & 72.84\textsubscript{±1.14} & 1.00  \\
                      & Uncertainty & 70.78\textsubscript{±0.59} & 14.58 \\
                      & Coreset     & 76.26\textsubscript{±0.48} & 0.20  \\
                      & Chameleon   & 72.97\textsubscript{±1.72} & 0.86  \\
                      & MOSAIC      & 77.38\textsubscript{±1.58} & 0.15  \\ \midrule
\multirow{5}{*}{500}  & Random      & 74.19\textsubscript{±1.05} & 1.00  \\
                      & Uncertainty & 69.77\textsubscript{±0.48} & 10.68 \\
                      & Coreset     & 78.12\textsubscript{±0.87} & 0.26  \\
                      & Chameleon   & 75.98\textsubscript{±0.06} & 0.70  \\
                      & MOSAIC      & 79.38\textsubscript{±1.05} & 0.20  \\ \midrule
\multirow{5}{*}{1000} & Random      & 75.84\textsubscript{±0.9}  & 1.00  \\
                      & Uncertainty & 71.12\textsubscript{±0.38} & NA    \\
                      & Coreset     & 80.46\textsubscript{±0.02} & 0.28  \\
                      & Chameleon   & 79.08\textsubscript{±0.74} & 0.44  \\
                      & MOSAIC      & 81.68\textsubscript{±0.52} & 0.19  \\ \midrule
\multirow{5}{*}{2000} & Random      & 78.39\textsubscript{±0.12} & 1.00  \\
                      & Uncertainty & 69.94\textsubscript{±1.4}  & NA    \\
                      & Coreset     & 81.37\textsubscript{±0.13} & 0.28  \\
                      & Chameleon   & 81.35\textsubscript{±0.39} & 0.44  \\
                      & MOSAIC      & 82.78\textsubscript{±0.41} & 0.19  \\ \midrule
\multirow{5}{*}{4000} & Random      & 80.38\textsubscript{±0.55} & 1.00  \\
                      & Uncertainty & 73.46\textsubscript{±0.19} & NA    \\
                      & Coreset     & 83.63\textsubscript{±0.36} & 0.25  \\
                      & Chameleon   & 82.92\textsubscript{±0.13} & 0.39  \\
                      & MOSAIC      & 84.25\textsubscript{±0.14} & 0.18  \\ \midrule
\multirow{5}{*}{8000} & Random      & 82.32\textsubscript{±0.54} & 1.00  \\
                      & Uncertainty & 75.63\textsubscript{±0.19} & NA    \\
                      & Coreset     & 84.49\textsubscript{±0.02} & 0.35  \\
                      & Chameleon   & 84.43\textsubscript{±0.01}  & 0.40  \\
                      & MOSAIC      & 85.02\textsubscript{±0.18} & 0.20  \\ \bottomrule
\end{tabular}
}
\end{table}

\begin{table}[h]
\centering
\caption{Navtrain validation EPDMS and BRMR results.}
\label{x:tab:main_navtrain}
\resizebox{0.8\linewidth}{!}{
\begin{tabular}{@{}cccc@{}}
\toprule
Budget                & Method      & EPDMS           & SRR  \\ \midrule
\multirow{5}{*}{100}  & Random      & 84.66\textsubscript{±0.6}  & 1.00 \\
                      & Uncertainty & 84.5\textsubscript{±0.48}  & 1.47 \\
                      & Coreset     & 85.29\textsubscript{±0.47} & 0.53 \\
                      & Chameleon   & 84.57\textsubscript{±0.18} & 1.07 \\
                      & MOSAIC    & 86.29\textsubscript{±0.43} & 0.30 \\ \midrule
\multirow{5}{*}{200}  & Random      & 85.45\textsubscript{±0.09} & 1.00 \\
                      & Uncertainty & 84.84\textsubscript{±0.54} & 1.50 \\
                      & Coreset     & 86.12\textsubscript{±0.31} & 0.60 \\
                      & Chameleon   & 86.04\textsubscript{±0.3}  & 0.80 \\
                      & MOSAIC     & 87.04\textsubscript{±0.37} & 0.32 \\ \midrule
\multirow{5}{*}{400}  & Random      & 86.69\textsubscript{±0.2}  & 1.00 \\
                      & Uncertainty & 86.07\textsubscript{±0.75} & 2.00 \\
                      & Coreset     & 87.09\textsubscript{±0.29} & 0.79 \\
                      & Chameleon   & 87.04\textsubscript{±0.6}  & 0.82 \\
                      & MOSAIC     & 88.21\textsubscript{±0.03} & 0.38 \\ \midrule
\multirow{5}{*}{800}  & Random      & 87.41\textsubscript{±0.37} & 1.00 \\
                      & Uncertainty & 86.69\textsubscript{±0.34} & 1.69 \\
                      & Coreset     & 88.48\textsubscript{±0.12} & 0.62 \\
                      & Chameleon   & 88.33\textsubscript{±0.23} & 0.64 \\
                      & MOSAIC    & 89.1\textsubscript{±0.12}  & 0.33 \\ \midrule
\multirow{5}{*}{1600} & Random      & 88.62\textsubscript{±0.22} & 1.00 \\
                      & Uncertainty & 87.75\textsubscript{±0.37} & 1.36 \\
                      & Coreset     & 89.3\textsubscript{±0.19}  & 0.58 \\
                      & Chameleon   & 89.5\textsubscript{±0.2}   & 0.62 \\
                      & MOSAIC     & 90.18\textsubscript{±0.25} & 0.37 \\ \midrule
\multirow{5}{*}{2400} & Random      & 89.42\textsubscript{±0.03} & 1.00 \\
                      & Uncertainty & 88.95\textsubscript{±0.15} & 1.00 \\
                      & Coreset     & 89.75\textsubscript{±0.02} & 0.76 \\
                      & Chameleon   & 90.05\textsubscript{±0.08} & 0.64 \\
                      & MOSAIC     & 90.31\textsubscript{±0.03} & 0.43 \\ \bottomrule
\end{tabular}
}
\end{table}

\begin{table}[]
\caption{Navtrain validation EPDMS and BRMR results under caption-based clustering.}
\label{x:tab:main_navtrain_caption}
\centering
\resizebox{0.8\linewidth}{!}{
\begin{tabular}{@{}cccc@{}}
\toprule
Budget                & Method      & EPDMS                      & SRR  \\ \midrule
\multirow{5}{*}{100}  & Random      & 84.66\textsubscript{±0.6}  & 1.00 \\
                      & Uncertainty & 84.5\textsubscript{±0.48}  & 1.47 \\
                      & Coreset     & 85.29\textsubscript{±0.47} & 0.53 \\
                      & Chameleon   & 84.35\textsubscript{±0.47} & 1.30 \\
                      & MOSAIC      & 85.85\textsubscript{±0.41} & 0.37 \\ \midrule
\multirow{5}{*}{200}  & Random      & 85.45\textsubscript{±0.09} & 1.00 \\
                      & Uncertainty & 84.84\textsubscript{±0.54} & 1.50 \\
                      & Coreset     & 86.12\textsubscript{±0.31} & 0.60 \\
                      & Chameleon   & 85.39\textsubscript{±0.02} & 2.88 \\
                      & MOSAIC      & 86.75\textsubscript{±0.17} & 0.40 \\ \midrule
\multirow{5}{*}{400}  & Random      & 86.69\textsubscript{±0.2}  & 1.00 \\
                      & Uncertainty & 86.07\textsubscript{±0.75} & 2.00 \\
                      & Coreset     & 87.09\textsubscript{±0.29} & 0.79 \\
                      & Chameleon   & 84.95\textsubscript{±0.45} & 3.32 \\
                      & MOSAIC      & 88.11\textsubscript{±0.05} & 0.48 \\ \midrule
\multirow{5}{*}{800}  & Random      & 87.41\textsubscript{±0.37} & 1.00 \\
                      & Uncertainty & 86.69\textsubscript{±0.34} & 1.69 \\
                      & Coreset     & 88.48\textsubscript{±0.12} & 0.62 \\
                      & Chameleon   & 86.1\textsubscript{±0.55}  & 2.68 \\
                      & MOSAIC      & 88.99\textsubscript{±0.09} & 0.37 \\ \midrule
\multirow{5}{*}{1600} & Random      & 88.62\textsubscript{±0.22} & 1.00 \\
                      & Uncertainty & 87.75\textsubscript{±0.37} & 1.36 \\
                      & Coreset     & 89.3\textsubscript{±0.19}  & 0.58 \\
                      & Chameleon   & 86.99\textsubscript{±0.57} & 1.50 \\
                      & MOSAIC      & 89.98\textsubscript{±0.13} & 0.39 \\ \midrule
\multirow{5}{*}{2400} & Random      & 89.42\textsubscript{±0.03} & 1.00 \\
                      & Uncertainty & 88.95\textsubscript{±0.15} & 1.00 \\
                      & Coreset     & 89.75\textsubscript{±0.02} & 0.76 \\
                      & Chameleon   & 87.62\textsubscript{±0.28} & 1.00 \\
                      & MOSAIC      & 90.37\textsubscript{±0.2}  & 0.48 \\ \bottomrule
\end{tabular}
}
\end{table}

\begin{table*}[h!]
\centering
\caption{Breakdown of the nine EPDMS rule-compliance metrics for the base model and the models trained with data selected by various strategies at all budgets, shown for the OpenScene experiment.}
\label{x:tab:breakdown_openscene}
\resizebox{1.0\linewidth}{!}{
\begin{tabular}{@{}cccccccccccc@{}}
\toprule
\multicolumn{2}{c}{Setting}         & NC                         & DAC                        & DDC                        & TLC                        & EP                         & TTC                        & LK                         & HC                         & EC                         & EPDMS                      \\ \midrule
\multicolumn{2}{c}{Base}            & 94.05                      & 83.9                       & 96.28                      & 99.6                       & 85.96                      & 92.95                      & 93.26                      & 98.25                      & 81.88                      & 72.0                       \\ \midrule
\multirow{5}{*}{250}  & Random      & 94.27\textsubscript{±0.60} & 84.63\textsubscript{±1.46} & 97.38\textsubscript{±0.23} & 99.66\textsubscript{±0.04} & 85.18\textsubscript{±1.02} & 93.23\textsubscript{±0.64} & 93.33\textsubscript{±0.56} & 98.26\textsubscript{±0.01} & 82.66\textsubscript{±0.76} & 72.84\textsubscript{±1.14} \\
                      & Uncertainty & 93.97\textsubscript{±0.44} & 82.49\textsubscript{±0.30} & 96.78\textsubscript{±0.44} & 99.66\textsubscript{±0.02} & 85.18\textsubscript{±0.81} & 92.98\textsubscript{±0.42} & 93.18\textsubscript{±0.66} & 98.23\textsubscript{±0.08} & 82.15\textsubscript{±0.27} & 70.78\textsubscript{±0.59} \\
                      & Coreset     & 95.11\textsubscript{±0.47} & 87.66\textsubscript{±0.61} & 98.38\textsubscript{±0.21} & 99.67\textsubscript{±0.04} & 86.09\textsubscript{±1.13} & 94.08\textsubscript{±0.84} & 94.47\textsubscript{±0.20} & 98.31\textsubscript{±0.05} & 83.38\textsubscript{±0.74} & 76.26\textsubscript{±0.48} \\
                      & Chameleon   & 94.02\textsubscript{±1.25} & 84.30\textsubscript{±1.18} & 97.48\textsubscript{±0.71} & 99.58\textsubscript{±0.06} & 87.48\textsubscript{±1.41} & 92.69\textsubscript{±1.23} & 93.43\textsubscript{±0.04} & 98.26\textsubscript{±0.01} & 83.15\textsubscript{±1.80} & 72.97\textsubscript{±1.72} \\
                      & MOSAIC      & 94.89\textsubscript{±0.74} & 88.76\textsubscript{±1.17} & 98.54\textsubscript{±0.43} & 99.61\textsubscript{±0.04} & 86.50\textsubscript{±1.03} & 93.93\textsubscript{±0.88} & 94.88\textsubscript{±0.14} & 98.26\textsubscript{±0.03} & 83.77\textsubscript{±0.67} & 77.38\textsubscript{±1.58} \\ \midrule
\multirow{5}{*}{500}  & Random      & 94.65\textsubscript{±0.21} & 85.72\textsubscript{±0.88} & 97.87\textsubscript{±0.44} & 99.64\textsubscript{±0.06} & 85.53\textsubscript{±0.22} & 93.51\textsubscript{±0.32} & 93.73\textsubscript{±0.24} & 98.27\textsubscript{±0.05} & 83.26\textsubscript{±0.14} & 74.19\textsubscript{±1.05} \\
                      & Uncertainty & 93.32\textsubscript{±0.47} & 82.26\textsubscript{±0.40} & 96.09\textsubscript{±0.52} & 99.60\textsubscript{±0.08} & 84.51\textsubscript{±0.43} & 92.23\textsubscript{±0.73} & 92.38\textsubscript{±0.56} & 98.30\textsubscript{±0.01} & 82.85\textsubscript{±1.07} & 69.77\textsubscript{±0.48} \\
                      & Coreset     & 95.56\textsubscript{±0.78} & 88.96\textsubscript{±0.57} & 98.95\textsubscript{±0.09} & 99.71\textsubscript{±0.07} & 86.21\textsubscript{±0.99} & 94.69\textsubscript{±0.79} & 95.14\textsubscript{±0.13} & 98.31\textsubscript{±0.03} & 84.24\textsubscript{±0.34} & 78.12\textsubscript{±0.87} \\
                      & Chameleon   & 95.00\textsubscript{±0.58} & 87.11\textsubscript{±0.09} & 98.16\textsubscript{±0.02} & 99.67\textsubscript{±0.16} & 86.67\textsubscript{±2.25} & 94.22\textsubscript{±0.45} & 94.20\textsubscript{±0.44} & 98.30\textsubscript{±0.01} & 83.69\textsubscript{±0.24} & 75.98\textsubscript{±0.06} \\
                      & MOSAIC      & 95.57\textsubscript{±1.05} & 90.54\textsubscript{±0.45} & 98.83\textsubscript{±0.29} & 99.67\textsubscript{±0.09} & 86.08\textsubscript{±1.79} & 94.85\textsubscript{±1.23} & 95.68\textsubscript{±0.27} & 98.25\textsubscript{±0.04} & 83.80\textsubscript{±0.16} & 79.38\textsubscript{±1.05} \\ \midrule
\multirow{5}{*}{1000} & Random      & 95.21\textsubscript{±0.58} & 87.15\textsubscript{±1.44} & 98.26\textsubscript{±0.39} & 99.72\textsubscript{±0.07} & 85.56\textsubscript{±0.96} & 94.35\textsubscript{±0.60} & 94.50\textsubscript{±0.66} & 98.31\textsubscript{±0.03} & 82.50\textsubscript{±0.52} & 75.84\textsubscript{±0.90} \\
                      & Uncertainty & 94.04\textsubscript{±0.70} & 83.77\textsubscript{±0.02} & 96.96\textsubscript{±0.08} & 99.70\textsubscript{±0.08} & 83.11\textsubscript{±0.66} & 93.21\textsubscript{±1.00} & 92.87\textsubscript{±0.14} & 98.32\textsubscript{±0.02} & 81.91\textsubscript{±0.75} & 71.12\textsubscript{±0.38} \\
                      & Coreset     & 95.93\textsubscript{±0.24} & 91.05\textsubscript{±0.26} & 99.28\textsubscript{±0.11} & 99.71\textsubscript{±0.04} & 86.39\textsubscript{±0.48} & 95.01\textsubscript{±0.21} & 95.75\textsubscript{±0.08} & 98.28\textsubscript{±0.03} & 84.58\textsubscript{±0.42} & 80.46\textsubscript{±0.02} \\
                      & Chameleon   & 95.89\textsubscript{±0.19} & 89.57\textsubscript{±0.68} & 98.94\textsubscript{±0.16} & 99.71\textsubscript{±0.07} & 86.39\textsubscript{±0.51} & 95.06\textsubscript{±0.26} & 95.44\textsubscript{±0.27} & 98.29\textsubscript{±0.01} & 84.23\textsubscript{±0.74} & 79.08\textsubscript{±0.74} \\
                      & MOSAIC      & 96.00\textsubscript{±0.22} & 92.20\textsubscript{±0.48} & 99.33\textsubscript{±0.07} & 99.67\textsubscript{±0.05} & 86.63\textsubscript{±0.41} & 95.24\textsubscript{±0.24} & 96.17\textsubscript{±0.22} & 98.28\textsubscript{±0.03} & 84.33\textsubscript{±0.30} & 81.68\textsubscript{±0.52} \\ \midrule
\multirow{5}{*}{2000} & Random      & 95.58\textsubscript{±0.54} & 89.26\textsubscript{±0.64} & 98.67\textsubscript{±0.18} & 99.70\textsubscript{±0.12} & 86.44\textsubscript{±0.42} & 94.88\textsubscript{±0.61} & 95.26\textsubscript{±0.18} & 98.30\textsubscript{±0.00} & 83.96\textsubscript{±0.96} & 78.39\textsubscript{±0.12} \\
                      & Uncertainty & 93.14\textsubscript{±0.79} & 82.66\textsubscript{±1.12} & 96.64\textsubscript{±0.64} & 99.53\textsubscript{±0.09} & 84.52\textsubscript{±1.23} & 92.19\textsubscript{±1.22} & 93.22\textsubscript{±0.29} & 98.28\textsubscript{±0.03} & 80.98\textsubscript{±1.30} & 69.94\textsubscript{±1.40} \\
                      & Coreset     & 95.89\textsubscript{±0.22} & 91.77\textsubscript{±0.14} & 99.44\textsubscript{±0.06} & 99.66\textsubscript{±0.04} & 87.39\textsubscript{±0.05} & 94.98\textsubscript{±0.18} & 95.99\textsubscript{±0.47} & 98.29\textsubscript{±0.00} & 85.55\textsubscript{±0.19} & 81.37\textsubscript{±0.13} \\
                      & Chameleon   & 96.38\textsubscript{±0.25} & 91.31\textsubscript{±0.26} & 99.15\textsubscript{±0.03} & 99.71\textsubscript{±0.05} & 86.55\textsubscript{±0.40} & 95.60\textsubscript{±0.29} & 95.99\textsubscript{±0.15} & 98.34\textsubscript{±0.01} & 85.04\textsubscript{±0.15} & 81.35\textsubscript{±0.39} \\
                      & MOSAIC      & 96.90\textsubscript{±0.38} & 92.29\textsubscript{±0.36} & 99.48\textsubscript{±0.05} & 99.73\textsubscript{±0.01} & 86.61\textsubscript{±0.73} & 96.16\textsubscript{±0.26} & 96.34\textsubscript{±0.06} & 98.28\textsubscript{±0.05} & 84.69\textsubscript{±0.01} & 82.78\textsubscript{±0.41} \\ \midrule
\multirow{5}{*}{4000} & Random      & 96.32\textsubscript{±0.59} & 90.53\textsubscript{±0.06} & 99.06\textsubscript{±0.07} & 99.79\textsubscript{±0.05} & 86.36\textsubscript{±0.48} & 95.66\textsubscript{±0.52} & 95.68\textsubscript{±0.09} & 98.30\textsubscript{±0.01} & 84.46\textsubscript{±0.14} & 80.38\textsubscript{±0.55} \\
                      & Uncertainty & 94.67\textsubscript{±0.28} & 85.11\textsubscript{±0.51} & 97.15\textsubscript{±0.54} & 99.71\textsubscript{±0.04} & 84.26\textsubscript{±0.69} & 93.72\textsubscript{±0.40} & 93.26\textsubscript{±0.09} & 98.28\textsubscript{±0.02} & 81.34\textsubscript{±1.06} & 73.46\textsubscript{±0.19} \\
                      & Coreset     & 97.11\textsubscript{±0.18} & 92.93\textsubscript{±0.60} & 99.44\textsubscript{±0.06} & 99.82\textsubscript{±0.02} & 86.65\textsubscript{±0.55} & 96.42\textsubscript{±0.19} & 96.66\textsubscript{±0.30} & 98.16\textsubscript{±0.12} & 85.10\textsubscript{±0.06} & 83.63\textsubscript{±0.36} \\
                      & Chameleon   & 96.76\textsubscript{±0.24} & 92.32\textsubscript{±0.02} & 99.51\textsubscript{±0.01} & 99.77\textsubscript{±0.01} & 86.98\textsubscript{±0.17} & 95.91\textsubscript{±0.31} & 96.49\textsubscript{±0.12} & 98.32\textsubscript{±0.01} & 85.51\textsubscript{±0.11} & 82.92\textsubscript{±0.13} \\
                      & MOSAIC      & 96.97\textsubscript{±0.32} & 93.59\textsubscript{±0.11} & 99.59\textsubscript{±0.04} & 99.80\textsubscript{±0.01} & 87.14\textsubscript{±0.98} & 96.18\textsubscript{±0.45} & 96.62\textsubscript{±0.08} & 98.28\textsubscript{±0.01} & 85.06\textsubscript{±0.34} & 84.25\textsubscript{±0.14} \\ \midrule
\multirow{5}{*}{8000} & Random      & 96.79\textsubscript{±0.21} & 91.88\textsubscript{±0.34} & 99.23\textsubscript{±0.11} & 99.79\textsubscript{±0.03} & 87.19\textsubscript{±0.05} & 95.93\textsubscript{±0.15} & 96.19\textsubscript{±0.10} & 98.28\textsubscript{±0.03} & 84.97\textsubscript{±0.19} & 82.32\textsubscript{±0.54} \\
                      & Uncertainty & 95.62\textsubscript{±0.38} & 86.48\textsubscript{±0.06} & 97.62\textsubscript{±0.01} & 99.71\textsubscript{±0.02} & 84.92\textsubscript{±0.25} & 94.80\textsubscript{±0.28} & 94.34\textsubscript{±0.27} & 98.32\textsubscript{±0.02} & 81.62\textsubscript{±0.09} & 75.63\textsubscript{±0.19} \\
                      & Coreset     & 97.39\textsubscript{±0.15} & 93.51\textsubscript{±0.18} & 99.55\textsubscript{±0.07} & 99.81\textsubscript{±0.03} & 87.07\textsubscript{±0.39} & 96.64\textsubscript{±0.12} & 96.78\textsubscript{±0.06} & 98.28\textsubscript{±0.03} & 85.51\textsubscript{±0.15} & 84.49\textsubscript{±0.02} \\
                      & Chameleon   & 97.33\textsubscript{±0.39} & 93.36\textsubscript{±0.14} & 99.61\textsubscript{±0.01} & 99.82\textsubscript{±0.01} & 87.34\textsubscript{±0.61} & 96.42\textsubscript{±0.50} & 96.90\textsubscript{±0.17} & 98.29\textsubscript{±0.02} & 85.51\textsubscript{±0.12} & 84.43\textsubscript{±0.00} \\
                      & MOSAIC      & 97.55\textsubscript{±0.13} & 93.84\textsubscript{±0.00} & 99.53\textsubscript{±0.18} & 99.84\textsubscript{±0.03} & 87.19\textsubscript{±0.24} & 96.79\textsubscript{±0.07} & 97.10\textsubscript{±0.07} & 98.29\textsubscript{±0.02} & 85.25\textsubscript{±0.22} & 85.02\textsubscript{±0.18} \\ \bottomrule
\end{tabular}
}
\end{table*}

\begin{table*}[h!]
\caption{Breakdown of the nine EPDMS rule-compliance metrics for the base model and the models trained with data selected by various strategies at all budgets, shown for the Navtrain experiment.}
\label{x:tab:breakdown_navtrain}
\resizebox{1.0\linewidth}{!}{
\begin{tabular}{@{}cccccccccccc@{}}
\toprule
\multicolumn{2}{c}{Setting}         & NC                         & DAC                        & DDC                        & TLC                        & EP                         & TTC                        & LK                         & HC                         & EC                         & EPDMS                      \\ \midrule
\multicolumn{2}{c}{Base}            & 95.3                       & 95.94                      & 99.09                      & 99.6                       & 88.09                      & 94.55                      & 94.49                      & 98.25                      & 82.39                      & 83.97                      \\ \midrule
\multirow{5}{*}{100}  & Random      & 95.43\textsubscript{±0.84} & 96.41\textsubscript{±0.20} & 98.98\textsubscript{±0.07} & 99.54\textsubscript{±0.15} & 88.68\textsubscript{±0.63} & 94.69\textsubscript{±0.89} & 94.82\textsubscript{±0.34} & 98.27\textsubscript{±0.04} & 82.81\textsubscript{±0.64} & 84.66\textsubscript{±0.60} \\
                      & Uncertainty & 95.68\textsubscript{±0.33} & 96.23\textsubscript{±0.38} & 98.91\textsubscript{±0.12} & 99.51\textsubscript{±0.06} & 88.21\textsubscript{±0.22} & 94.77\textsubscript{±0.36} & 94.88\textsubscript{±0.13} & 98.27\textsubscript{±0.04} & 83.50\textsubscript{±0.32} & 84.50\textsubscript{±0.48} \\
                      & Coreset     & 95.63\textsubscript{±0.41} & 96.88\textsubscript{±0.33} & 99.13\textsubscript{±0.09} & 99.56\textsubscript{±0.03} & 88.39\textsubscript{±0.65} & 94.75\textsubscript{±0.51} & 94.97\textsubscript{±0.34} & 98.25\textsubscript{±0.03} & 82.94\textsubscript{±0.20} & 85.29\textsubscript{±0.47} \\
                      & Chameleon   & 95.14\textsubscript{±0.20} & 96.50\textsubscript{±0.19} & 99.17\textsubscript{±0.02} & 99.53\textsubscript{±0.02} & 88.80\textsubscript{±0.18} & 94.35\textsubscript{±0.11} & 95.10\textsubscript{±0.12} & 98.25\textsubscript{±0.04} & 82.93\textsubscript{±0.75} & 84.57\textsubscript{±0.18} \\
                      & MOSAIC      & 96.75\textsubscript{±0.28} & 97.06\textsubscript{±0.09} & 99.03\textsubscript{±0.03} & 99.60\textsubscript{±0.03} & 87.74\textsubscript{±0.28} & 96.09\textsubscript{±0.35} & 94.92\textsubscript{±0.32} & 98.27\textsubscript{±0.02} & 82.80\textsubscript{±0.62} & 86.29\textsubscript{±0.43} \\ \midrule
\multirow{5}{*}{200}  & Random      & 95.90\textsubscript{±0.35} & 96.58\textsubscript{±0.23} & 99.11\textsubscript{±0.08} & 99.65\textsubscript{±0.02} & 88.75\textsubscript{±0.19} & 95.08\textsubscript{±0.33} & 95.14\textsubscript{±0.29} & 98.27\textsubscript{±0.04} & 83.14\textsubscript{±0.11} & 85.45\textsubscript{±0.09} \\
                      & Uncertainty & 95.61\textsubscript{±0.66} & 96.53\textsubscript{±0.38} & 98.96\textsubscript{±0.18} & 99.57\textsubscript{±0.09} & 88.51\textsubscript{±0.11} & 94.74\textsubscript{±0.59} & 94.88\textsubscript{±0.28} & 98.28\textsubscript{±0.03} & 83.34\textsubscript{±0.34} & 84.84\textsubscript{±0.54} \\
                      & Coreset     & 96.19\textsubscript{±0.49} & 97.05\textsubscript{±0.11} & 99.13\textsubscript{±0.06} & 99.60\textsubscript{±0.04} & 88.68\textsubscript{±0.13} & 95.39\textsubscript{±0.44} & 95.17\textsubscript{±0.11} & 98.29\textsubscript{±0.01} & 83.45\textsubscript{±0.55} & 86.12\textsubscript{±0.31} \\
                      & Chameleon   & 96.12\textsubscript{±0.52} & 96.76\textsubscript{±0.34} & 99.33\textsubscript{±0.18} & 99.60\textsubscript{±0.10} & 88.74\textsubscript{±0.54} & 95.38\textsubscript{±0.71} & 95.41\textsubscript{±0.19} & 98.30\textsubscript{±0.02} & 83.73\textsubscript{±0.13} & 86.04\textsubscript{±0.30} \\
                      & MOSAIC    & 96.83\textsubscript{±0.31} & 97.51\textsubscript{±0.18} & 99.24\textsubscript{±0.06} & 99.61\textsubscript{±0.01} & 88.20\textsubscript{±0.13} & 96.16\textsubscript{±0.29} & 95.36\textsubscript{±0.18} & 98.26\textsubscript{±0.02} & 82.63\textsubscript{±0.35} & 87.04\textsubscript{±0.37} \\ \midrule
\multirow{5}{*}{400}  & Random      & 96.71\textsubscript{±0.25} & 96.91\textsubscript{±0.20} & 99.18\textsubscript{±0.09} & 99.71\textsubscript{±0.01} & 88.75\textsubscript{±0.15} & 96.02\textsubscript{±0.23} & 95.76\textsubscript{±0.16} & 98.30\textsubscript{±0.01} & 82.96\textsubscript{±0.10} & 86.69\textsubscript{±0.20} \\
                      & Uncertainty & 96.39\textsubscript{±0.60} & 96.97\textsubscript{±0.38} & 99.00\textsubscript{±0.08} & 99.65\textsubscript{±0.01} & 88.22\textsubscript{±0.42} & 95.55\textsubscript{±0.66} & 94.98\textsubscript{±0.22} & 98.25\textsubscript{±0.02} & 83.64\textsubscript{±0.16} & 86.07\textsubscript{±0.75} \\
                      & Coreset     & 96.73\textsubscript{±0.24} & 97.27\textsubscript{±0.17} & 99.36\textsubscript{±0.02} & 99.64\textsubscript{±0.02} & 88.80\textsubscript{±0.11} & 95.95\textsubscript{±0.26} & 95.81\textsubscript{±0.23} & 98.29\textsubscript{±0.03} & 83.48\textsubscript{±0.46} & 87.09\textsubscript{±0.29} \\
                      & Chameleon   & 96.33\textsubscript{±0.36} & 97.55\textsubscript{±0.20} & 99.37\textsubscript{±0.07} & 99.63\textsubscript{±0.03} & 88.97\textsubscript{±0.42} & 95.59\textsubscript{±0.38} & 95.87\textsubscript{±0.20} & 98.30\textsubscript{±0.01} & 83.10\textsubscript{±0.35} & 87.04\textsubscript{±0.60} \\
                      & MOSAIC    & 97.75\textsubscript{±0.08} & 97.79\textsubscript{±0.11} & 99.42\textsubscript{±0.06} & 99.72\textsubscript{±0.04} & 87.62\textsubscript{±0.11} & 97.17\textsubscript{±0.09} & 95.54\textsubscript{±0.08} & 98.24\textsubscript{±0.01} & 82.81\textsubscript{±0.27} & 88.21\textsubscript{±0.03} \\ \midrule
\multirow{5}{*}{800}  & Random      & 96.94\textsubscript{±0.35} & 97.15\textsubscript{±0.36} & 99.35\textsubscript{±0.12} & 99.69\textsubscript{±0.05} & 89.16\textsubscript{±0.06} & 96.22\textsubscript{±0.41} & 96.28\textsubscript{±0.45} & 98.29\textsubscript{±0.03} & 83.63\textsubscript{±0.02} & 87.41\textsubscript{±0.37} \\
                      & Uncertainty & 96.98\textsubscript{±0.40} & 96.88\textsubscript{±0.16} & 99.13\textsubscript{±0.11} & 99.69\textsubscript{±0.07} & 88.31\textsubscript{±0.45} & 96.22\textsubscript{±0.32} & 95.42\textsubscript{±0.15} & 98.28\textsubscript{±0.03} & 82.95\textsubscript{±0.31} & 86.69\textsubscript{±0.34} \\
                      & Coreset     & 97.21\textsubscript{±0.12} & 98.06\textsubscript{±0.23} & 99.49\textsubscript{±0.06} & 99.67\textsubscript{±0.05} & 88.84\textsubscript{±0.08} & 96.62\textsubscript{±0.13} & 96.22\textsubscript{±0.19} & 98.30\textsubscript{±0.03} & 83.67\textsubscript{±0.27} & 88.48\textsubscript{±0.12} \\
                      & Chameleon   & 97.07\textsubscript{±0.17} & 97.97\textsubscript{±0.18} & 99.48\textsubscript{±0.08} & 99.68\textsubscript{±0.03} & 88.99\textsubscript{±0.50} & 96.57\textsubscript{±0.19} & 96.38\textsubscript{±0.18} & 98.29\textsubscript{±0.03} & 83.28\textsubscript{±0.22} & 88.33\textsubscript{±0.23} \\
                      & MOSAIC    & 97.65\textsubscript{±0.14} & 98.33\textsubscript{±0.06} & 99.54\textsubscript{±0.05} & 99.73\textsubscript{±0.05} & 88.68\textsubscript{±0.44} & 97.03\textsubscript{±0.16} & 96.19\textsubscript{±0.14} & 98.26\textsubscript{±0.02} & 82.93\textsubscript{±0.48} & 89.10\textsubscript{±0.12} \\ \midrule
\multirow{5}{*}{1600} & Random      & 97.17\textsubscript{±0.07} & 98.19\textsubscript{±0.43} & 99.42\textsubscript{±0.05} & 99.69\textsubscript{±0.02} & 89.36\textsubscript{±0.12} & 96.50\textsubscript{±0.14} & 96.45\textsubscript{±0.25} & 98.31\textsubscript{±0.03} & 83.17\textsubscript{±0.76} & 88.62\textsubscript{±0.22} \\
                      & Uncertainty & 96.92\textsubscript{±0.38} & 97.66\textsubscript{±0.08} & 99.22\textsubscript{±0.10} & 99.77\textsubscript{±0.02} & 89.02\textsubscript{±0.28} & 96.24\textsubscript{±0.40} & 96.10\textsubscript{±0.07} & 98.30\textsubscript{±0.01} & 82.92\textsubscript{±0.38} & 87.75\textsubscript{±0.37} \\
                      & Coreset     & 97.50\textsubscript{±0.10} & 98.31\textsubscript{±0.34} & 99.59\textsubscript{±0.03} & 99.72\textsubscript{±0.05} & 89.27\textsubscript{±0.21} & 96.86\textsubscript{±0.07} & 96.75\textsubscript{±0.22} & 98.30\textsubscript{±0.03} & 83.88\textsubscript{±0.50} & 89.30\textsubscript{±0.19} \\
                      & Chameleon   & 97.43\textsubscript{±0.22} & 98.46\textsubscript{±0.17} & 99.60\textsubscript{±0.05} & 99.75\textsubscript{±0.03} & 89.60\textsubscript{±0.19} & 96.83\textsubscript{±0.30} & 96.89\textsubscript{±0.07} & 98.30\textsubscript{±0.03} & 83.87\textsubscript{±0.34} & 89.50\textsubscript{±0.20} \\
                      & MOSAIC    & 98.04\textsubscript{±0.24} & 98.61\textsubscript{±0.32} & 99.63\textsubscript{±0.06} & 99.73\textsubscript{±0.02} & 89.28\textsubscript{±0.19} & 97.50\textsubscript{±0.32} & 97.07\textsubscript{±0.06} & 98.28\textsubscript{±0.04} & 83.70\textsubscript{±0.41} & 90.18\textsubscript{±0.25} \\ \midrule
\multirow{5}{*}{2400} & Random      & 97.56\textsubscript{±0.11} & 98.23\textsubscript{±0.12} & 99.56\textsubscript{±0.04} & 99.74\textsubscript{±0.00} & 89.57\textsubscript{±0.08} & 96.97\textsubscript{±0.09} & 96.95\textsubscript{±0.07} & 98.30\textsubscript{±0.01} & 83.95\textsubscript{±0.31} & 89.42\textsubscript{±0.03} \\
                      & Uncertainty & 97.62\textsubscript{±0.21} & 98.10\textsubscript{±0.17} & 99.36\textsubscript{±0.10} & 99.78\textsubscript{±0.02} & 89.19\textsubscript{±0.15} & 97.07\textsubscript{±0.22} & 96.65\textsubscript{±0.27} & 98.29\textsubscript{±0.05} & 82.53\textsubscript{±0.48} & 88.95\textsubscript{±0.15} \\
                      & Coreset     & 97.59\textsubscript{±0.02} & 98.53\textsubscript{±0.06} & 99.57\textsubscript{±0.04} & 99.67\textsubscript{±0.04} & 89.79\textsubscript{±0.24} & 97.17\textsubscript{±0.13} & 97.17\textsubscript{±0.10} & 98.31\textsubscript{±0.02} & 83.77\textsubscript{±0.54} & 89.75\textsubscript{±0.02} \\
                      & Chameleon   & 97.60\textsubscript{±0.16} & 98.71\textsubscript{±0.06} & 99.63\textsubscript{±0.04} & 99.77\textsubscript{±0.01} & 89.85\textsubscript{±0.06} & 97.18\textsubscript{±0.14} & 97.20\textsubscript{±0.09} & 98.28\textsubscript{±0.01} & 83.61\textsubscript{±0.51} & 90.05\textsubscript{±0.08} \\
                      & MOSAIC    & 98.02\textsubscript{±0.12} & 98.69\textsubscript{±0.05} & 99.66\textsubscript{±0.07} & 99.80\textsubscript{±0.06} & 89.19\textsubscript{±0.38} & 97.58\textsubscript{±0.10} & 97.22\textsubscript{±0.09} & 98.31\textsubscript{±0.00} & 83.56\textsubscript{±0.07} & 90.31\textsubscript{±0.03} \\ \bottomrule
\end{tabular}
}
\end{table*}

\begin{table*}[]
\caption{Breakdown of the nine EPDMS rule-compliance metrics for the base model and the models trained with data selected by various strategies at all budgets, shown for the Navtrain experiment when the clustering is performed on the clip captions.}
\label{x:tab:breakdown_navtrain_captions}
\resizebox{1.0\linewidth}{!}{
\begin{tabular}{@{}cccccccccccc@{}}
\toprule
\multicolumn{2}{c}{Setting}         & NC                         & DAC                        & DDC                        & TLC                        & EP                         & TTC                        & LK                         & HC                         & EC                         & EPDMS                      \\ \midrule
\multicolumn{2}{c}{Base}            & 95.3                       & 95.94                      & 99.09                      & 99.6                       & 88.09                      & 94.55                      & 94.49                      & 98.25                      & 82.39                      & 83.97                      \\ \midrule
\multirow{5}{*}{100}  & Random      & 95.43\textsubscript{±0.84} & 96.41\textsubscript{±0.20} & 98.98\textsubscript{±0.07} & 99.54\textsubscript{±0.15} & 88.68\textsubscript{±0.63} & 94.69\textsubscript{±0.89} & 94.82\textsubscript{±0.34} & 98.27\textsubscript{±0.04} & 82.81\textsubscript{±0.64} & 84.66\textsubscript{±0.60} \\
                      & Uncertainty & 95.68\textsubscript{±0.33} & 96.23\textsubscript{±0.38} & 98.91\textsubscript{±0.12} & 99.51\textsubscript{±0.06} & 88.21\textsubscript{±0.22} & 94.77\textsubscript{±0.36} & 94.88\textsubscript{±0.13} & 98.27\textsubscript{±0.04} & 83.50\textsubscript{±0.32} & 84.50\textsubscript{±0.48} \\
                      & Coreset     & 95.63\textsubscript{±0.41} & 96.88\textsubscript{±0.33} & 99.13\textsubscript{±0.09} & 99.56\textsubscript{±0.03} & 88.39\textsubscript{±0.65} & 94.75\textsubscript{±0.51} & 94.97\textsubscript{±0.34} & 98.25\textsubscript{±0.03} & 82.94\textsubscript{±0.20} & 85.29\textsubscript{±0.47} \\
                      & Chameleon   & 95.43\textsubscript{±0.61} & 96.14\textsubscript{±0.02} & 98.94\textsubscript{±0.12} & 99.56\textsubscript{±0.06} & 88.45\textsubscript{±0.26} & 94.52\textsubscript{±0.59} & 94.82\textsubscript{±0.07} & 98.28\textsubscript{±0.02} & 83.27\textsubscript{±0.49} & 84.35\textsubscript{±0.47} \\
                      & MOSAIC    & 96.53\textsubscript{±0.31} & 96.91\textsubscript{±0.30} & 99.03\textsubscript{±0.12} & 99.54\textsubscript{±0.06} & 87.62\textsubscript{±0.21} & 95.80\textsubscript{±0.32} & 94.85\textsubscript{±0.34} & 98.24\textsubscript{±0.02} & 82.66\textsubscript{±0.66} & 85.85\textsubscript{±0.41} \\ \midrule
\multirow{5}{*}{200}  & Random      & 95.90\textsubscript{±0.35} & 96.58\textsubscript{±0.23} & 99.11\textsubscript{±0.08} & 99.65\textsubscript{±0.02} & 88.75\textsubscript{±0.19} & 95.08\textsubscript{±0.33} & 95.14\textsubscript{±0.29} & 98.27\textsubscript{±0.04} & 83.14\textsubscript{±0.11} & 85.45\textsubscript{±0.09} \\
                      & Uncertainty & 95.61\textsubscript{±0.66} & 96.53\textsubscript{±0.38} & 98.96\textsubscript{±0.18} & 99.57\textsubscript{±0.09} & 88.51\textsubscript{±0.11} & 94.74\textsubscript{±0.59} & 94.88\textsubscript{±0.28} & 98.28\textsubscript{±0.03} & 83.34\textsubscript{±0.34} & 84.84\textsubscript{±0.54} \\
                      & Coreset     & 96.19\textsubscript{±0.49} & 97.05\textsubscript{±0.11} & 99.13\textsubscript{±0.06} & 99.60\textsubscript{±0.04} & 88.68\textsubscript{±0.13} & 95.39\textsubscript{±0.44} & 95.17\textsubscript{±0.11} & 98.29\textsubscript{±0.01} & 83.45\textsubscript{±0.55} & 86.12\textsubscript{±0.31} \\
                      & Chameleon   & 96.20\textsubscript{±0.11} & 96.58\textsubscript{±0.25} & 98.97\textsubscript{±0.21} & 99.63\textsubscript{±0.01} & 88.02\textsubscript{±0.19} & 95.42\textsubscript{±0.24} & 94.88\textsubscript{±0.23} & 98.30\textsubscript{±0.00} & 82.88\textsubscript{±1.50} & 85.39\textsubscript{±0.02} \\
                      & MOSAIC    & 97.07\textsubscript{±0.30} & 97.19\textsubscript{±0.25} & 99.08\textsubscript{±0.06} & 99.64\textsubscript{±0.03} & 87.79\textsubscript{±0.46} & 96.28\textsubscript{±0.36} & 94.99\textsubscript{±0.29} & 98.25\textsubscript{±0.01} & 82.92\textsubscript{±0.55} & 86.75\textsubscript{±0.17} \\ \midrule
\multirow{5}{*}{400}  & Random      & 96.71\textsubscript{±0.25} & 96.91\textsubscript{±0.20} & 99.18\textsubscript{±0.09} & 99.71\textsubscript{±0.01} & 88.75\textsubscript{±0.15} & 96.02\textsubscript{±0.23} & 95.76\textsubscript{±0.16} & 98.30\textsubscript{±0.01} & 82.96\textsubscript{±0.10} & 86.69\textsubscript{±0.20} \\
                      & Uncertainty & 96.39\textsubscript{±0.60} & 96.97\textsubscript{±0.38} & 99.00\textsubscript{±0.08} & 99.65\textsubscript{±0.01} & 88.22\textsubscript{±0.42} & 95.55\textsubscript{±0.66} & 94.98\textsubscript{±0.22} & 98.25\textsubscript{±0.02} & 83.64\textsubscript{±0.16} & 86.07\textsubscript{±0.75} \\
                      & Coreset     & 96.73\textsubscript{±0.24} & 97.27\textsubscript{±0.17} & 99.36\textsubscript{±0.02} & 99.64\textsubscript{±0.02} & 88.80\textsubscript{±0.11} & 95.95\textsubscript{±0.26} & 95.81\textsubscript{±0.23} & 98.29\textsubscript{±0.03} & 83.48\textsubscript{±0.46} & 87.09\textsubscript{±0.29} \\
                      & Chameleon   & 95.61\textsubscript{±0.32} & 96.50\textsubscript{±0.25} & 99.04\textsubscript{±0.11} & 99.59\textsubscript{±0.06} & 88.64\textsubscript{±0.56} & 94.84\textsubscript{±0.42} & 94.90\textsubscript{±0.13} & 98.29\textsubscript{±0.00} & 82.73\textsubscript{±0.34} & 84.95\textsubscript{±0.45} \\
                      & MOSAIC    & 97.36\textsubscript{±0.10} & 97.91\textsubscript{±0.05} & 99.33\textsubscript{±0.10} & 99.66\textsubscript{±0.02} & 88.37\textsubscript{±0.41} & 96.68\textsubscript{±0.11} & 95.43\textsubscript{±0.34} & 98.27\textsubscript{±0.03} & 83.00\textsubscript{±1.38} & 88.11\textsubscript{±0.05} \\ \midrule
\multirow{5}{*}{800}  & Random      & 96.94\textsubscript{±0.35} & 97.15\textsubscript{±0.36} & 99.35\textsubscript{±0.12} & 99.69\textsubscript{±0.05} & 89.16\textsubscript{±0.06} & 96.22\textsubscript{±0.41} & 96.28\textsubscript{±0.45} & 98.29\textsubscript{±0.03} & 83.63\textsubscript{±0.02} & 87.41\textsubscript{±0.37} \\
                      & Uncertainty & 96.98\textsubscript{±0.40} & 96.88\textsubscript{±0.16} & 99.13\textsubscript{±0.11} & 99.69\textsubscript{±0.07} & 88.31\textsubscript{±0.45} & 96.22\textsubscript{±0.32} & 95.42\textsubscript{±0.15} & 98.28\textsubscript{±0.03} & 82.95\textsubscript{±0.31} & 86.69\textsubscript{±0.34} \\
                      & Coreset     & 97.21\textsubscript{±0.12} & 98.06\textsubscript{±0.23} & 99.49\textsubscript{±0.06} & 99.67\textsubscript{±0.05} & 88.84\textsubscript{±0.08} & 96.62\textsubscript{±0.13} & 96.22\textsubscript{±0.19} & 98.30\textsubscript{±0.03} & 83.67\textsubscript{±0.27} & 88.48\textsubscript{±0.12} \\
                      & Chameleon   & 96.26\textsubscript{±0.49} & 96.83\textsubscript{±0.47} & 99.10\textsubscript{±0.05} & 99.71\textsubscript{±0.03} & 88.89\textsubscript{±0.48} & 95.53\textsubscript{±0.46} & 95.64\textsubscript{±0.21} & 98.28\textsubscript{±0.01} & 82.97\textsubscript{±0.19} & 86.10\textsubscript{±0.55} \\
                      & MOSAIC    & 97.92\textsubscript{±0.09} & 98.08\textsubscript{±0.17} & 99.50\textsubscript{±0.05} & 99.73\textsubscript{±0.01} & 88.20\textsubscript{±0.25} & 97.35\textsubscript{±0.14} & 96.12\textsubscript{±0.22} & 98.25\textsubscript{±0.04} & 83.00\textsubscript{±0.50} & 88.99\textsubscript{±0.09} \\ \midrule
\multirow{5}{*}{1600} & Random      & 97.17\textsubscript{±0.07} & 98.19\textsubscript{±0.43} & 99.42\textsubscript{±0.05} & 99.69\textsubscript{±0.02} & 89.36\textsubscript{±0.12} & 96.50\textsubscript{±0.14} & 96.45\textsubscript{±0.25} & 98.31\textsubscript{±0.03} & 83.17\textsubscript{±0.76} & 88.62\textsubscript{±0.22} \\
                      & Uncertainty & 96.92\textsubscript{±0.38} & 97.66\textsubscript{±0.08} & 99.22\textsubscript{±0.10} & 99.77\textsubscript{±0.02} & 89.02\textsubscript{±0.28} & 96.24\textsubscript{±0.40} & 96.10\textsubscript{±0.07} & 98.30\textsubscript{±0.01} & 82.92\textsubscript{±0.38} & 87.75\textsubscript{±0.37} \\
                      & Coreset     & 97.50\textsubscript{±0.10} & 98.31\textsubscript{±0.34} & 99.59\textsubscript{±0.03} & 99.72\textsubscript{±0.05} & 89.27\textsubscript{±0.21} & 96.86\textsubscript{±0.07} & 96.75\textsubscript{±0.22} & 98.30\textsubscript{±0.03} & 83.88\textsubscript{±0.50} & 89.30\textsubscript{±0.19} \\
                      & Chameleon   & 96.61\textsubscript{±0.30} & 97.22\textsubscript{±0.26} & 99.25\textsubscript{±0.11} & 99.72\textsubscript{±0.06} & 89.05\textsubscript{±0.24} & 96.02\textsubscript{±0.39} & 95.90\textsubscript{±0.18} & 98.32\textsubscript{±0.01} & 82.35\textsubscript{±0.27} & 86.99\textsubscript{±0.57} \\
                      & MOSAIC    & 97.98\textsubscript{±0.05} & 98.59\textsubscript{±0.12} & 99.60\textsubscript{±0.03} & 99.76\textsubscript{±0.01} & 89.03\textsubscript{±0.24} & 97.49\textsubscript{±0.11} & 97.02\textsubscript{±0.25} & 98.27\textsubscript{±0.03} & 83.61\textsubscript{±0.31} & 89.98\textsubscript{±0.13} \\ \midrule
\multirow{5}{*}{2400} & Random      & 97.56\textsubscript{±0.11} & 98.23\textsubscript{±0.12} & 99.56\textsubscript{±0.04} & 99.74\textsubscript{±0.00} & 89.57\textsubscript{±0.08} & 96.97\textsubscript{±0.09} & 96.95\textsubscript{±0.07} & 98.30\textsubscript{±0.01} & 83.95\textsubscript{±0.31} & 89.42\textsubscript{±0.03} \\
                      & Uncertainty & 97.62\textsubscript{±0.21} & 98.10\textsubscript{±0.17} & 99.36\textsubscript{±0.10} & 99.78\textsubscript{±0.02} & 89.19\textsubscript{±0.15} & 97.07\textsubscript{±0.22} & 96.65\textsubscript{±0.27} & 98.29\textsubscript{±0.05} & 82.53\textsubscript{±0.48} & 88.95\textsubscript{±0.15} \\
                      & Coreset     & 97.59\textsubscript{±0.02} & 98.53\textsubscript{±0.06} & 99.57\textsubscript{±0.04} & 99.67\textsubscript{±0.04} & 89.79\textsubscript{±0.24} & 97.17\textsubscript{±0.13} & 97.17\textsubscript{±0.10} & 98.31\textsubscript{±0.02} & 83.77\textsubscript{±0.54} & 89.75\textsubscript{±0.02} \\
                      & Chameleon   & 96.93\textsubscript{±0.30} & 97.50\textsubscript{±0.25} & 99.37\textsubscript{±0.05} & 99.75\textsubscript{±0.03} & 88.97\textsubscript{±0.44} & 96.37\textsubscript{±0.24} & 96.22\textsubscript{±0.18} & 98.31\textsubscript{±0.00} & 82.39\textsubscript{±0.45} & 87.62\textsubscript{±0.28} \\
                      & MOSAIC    & 98.03\textsubscript{±0.28} & 98.78\textsubscript{±0.15} & 99.62\textsubscript{±0.02} & 99.79\textsubscript{±0.07} & 89.26\textsubscript{±0.45} & 97.59\textsubscript{±0.27} & 96.97\textsubscript{±0.12} & 98.33\textsubscript{±0.03} & 84.02\textsubscript{±0.10} & 90.37\textsubscript{±0.20} \\ \bottomrule
\end{tabular}
}
\end{table*}

\end{document}